\documentclass[10pt,journal,compsoc]{IEEEtran}

\ifCLASSOPTIONcompsoc
  \usepackage[nocompress]{cite}
\else
  \usepackage{cite}
\fi

\ifCLASSINFOpdf
  \usepackage[pdftex]{graphicx}
\else
\fi

\usepackage{amsmath}

\usepackage{placeins}
\usepackage{stfloats}
\usepackage{url}

\usepackage{tikz}
\definecolor{gold}{rgb}{0.85,.66,0}
\definecolor{tiffani}{cmyk}{0.73,0,0.35,0}

\begin{document}

\title{Concepts and Experiments on Psychoanalysis Driven Computing}


\author{\IEEEauthorblockN{Minas Gadalla}\\
\IEEEauthorblockA{Department of Computer Engineering \& Informatics\\
University of Patras, Greece\\
email: gkantalla@ceid.upatras.gr}\\
\and
\IEEEauthorblockN{Sotiris Nikoletseas}\\
\IEEEauthorblockA{University of Patras and CTI, Greece\\
email: nikole@cti.gr}\\
\and
\IEEEauthorblockN{José Roberto de A. Amazonas}\\
\IEEEauthorblockA{Department of Computer Architecture\\
Technical University of Catalonia - UPC\\
Barcelona - Spain\\
\and
email: jose.roberto.amazonas@upc.edu}
\IEEEauthorblockN{José D. P. Rolim}\\
\IEEEauthorblockA{University of Geneva, Switzerland\\
email: jose.rolim@unige.ch}}


%


\maketitle

\begin{abstract}
This research investigates the effective incorporation of the human factor and user perception in text-based interactive media. In such contexts, the reliability of user texts is often compromised by behavioural and emotional dimensions. To this end, several attempts have been made in the state of the art, to introduce psychological approaches in such systems, including computational psycholinguistics, personality traits and cognitive psychology methods. 

In contrast, our method is fundamentally different since we employ a psychoanalysis-based approach; in particular, we use the notion of Lacanian discourse types, to capture and deeply understand real (possibly elusive) characteristics, qualities and contents of texts, and evaluate their reliability. As far as we know, this is the first time computational methods are systematically combined with psychoanalysis. We believe such psychoanalytic framework is fundamentally more effective than standard methods, since it addresses deeper, quite primitive elements of human personality, behaviour and expression which usually escape methods functioning at ``higher'', conscious layers. In fact, this research is a first attempt to form a new paradigm of psychoanalysis-driven interactive technologies, with broader impact and diverse applications.

To exemplify this generic approach, we apply it to the case-study of fake news detection; we first demonstrate certain limitations of the well-known Myers–Briggs Type Indicator (MBTI) personality type method, and then propose and evaluate our new method of analysing user texts and detecting fake news based on the Lacanian discourses psychoanalytic approach.
\end{abstract}


%
\IEEEpeerreviewmaketitle

\section{Introduction}
\label{sec:introduction}
User-related and generated data (for simplicity referred to as user data) constitute a core component for social applications based on interactive media technologies. However, in various contexts where user perception is more susceptible to emotional or some other form of bias, user data reliability is often compromised. 

A scenario that accurately exemplifies the impact of the lack of reliability of user data can be illustrated in disaster management. In a hypothetical event of a car accident on a frequented highway, where lots of other cars are passing by, we can imagine loads of social media posts describing details about it, before any truly reliable information (e.g. the police arriving) is conveyed. In such a case, if the high volume/uncertain quality user-based information could be instantly filtered, so that a reliable source which e.g. accurately reports the degree of passenger injury severity could be identified, the timely arrival of an ambulance could be of life-saving impact. Of course, the above is a non-trivial task as it implies mechanisms which are yet non-existent; it is certain, though, that an interdisciplinary approach is necessary to capture the diverse aspects of human social way of expressing perception in terms of data and model it in a formal way, in order to conceptualise such mechanisms and infer the reliability of the information and knowledge that can be acquired from such data.

Because of this lack of reliability associated with user data, interactive media technologies-based applications that depend on the characterisation and prediction of the information and knowledge acquired from such data may be severely impaired and are rarely adopted by the actors involved in real life scenarios. 

A relevant, recently established research topic is on detecting fake news. In particular, ``fake news detection'' is defined as the task of categorising news along a continuum of veracity, with an associated measure of certainty; veracity is compromised by the occurrence of intentional deceptions \cite{conroy2015}. The state-of-the-art methods for detecting the spread of fake news can be coarsely classified into two categories. The linguistic approaches are based on ``language leakages'' that take place when someone tries to conceal a lie \cite{Feng2013}. Here, certain verbal aspects are monitored, such as frequencies and patterns of pronouns, conjunctions, and negative emotion word usage; a task that is found very difficult to achieve. On the other hand, the network approaches are based on corresponding properties and behaviour of how the news is spread. Here, linked data and social network behaviour are studied \cite{Ciampaglia2015}. 

We conjecture that psychoanalysis theories may be used to provide the tools for a third methodology to be developed. One may choose to disseminate fake news for several reasons; e.g., due to being irrational or because there is something to gain. Independently of the different motives, certain text qualities characteristic of fake news can be captured by a psychoanalytic examination of the texts.

The overarching aim of this research work is to develop a radically new theoretical framework for interpreting user-generated data in the context of social interactions. In particular, by combining elements from two very divergent disciplines -- Computer Science and Psychoanalysis -- this work will develop the theoretical methods and tools for gaining a deep, holistic understanding of the behavioural context of individuals, groups and crowds from the data they generate. This work focus on improving and providing a fundamentally new perspective \textbf{in terms of the corresponding technologies}.

 In this context, this research has the ambitious goal of laying the foundations for a new paradigm of {\bf psychoanalysis-driven technologies}.  
 
 The specific contributions of this paper, within the broader aims stated previously, may be summarised as follows:
 \begin{itemize}
 	\item evaluation of the fake news detection accuracy method based on the personality traits concept demonstrating its limitations;
	
	\item proposal of a new method to classify user data based on the Lacanian discourses psychoanalytical concept;
	
	\item evaluation of the fake news detection accuracy based on the Lacanian discourses approach;
	
	\item definition of a framework and roadmap for the future development of psychoanalysis-driven computing.
 \end{itemize}

After this Introduction, Section \ref{sec:relatedwork} describes and compares related works published in the recent years, Section \ref{sec:psyapproaches} describes the personality traits concept and introduces the novel psychoanalysis-driven approach based on Lacanian discourses, Section \ref{sec:preliminaryevaluation} evaluates the potential of the adopted Psychological and Psychoanalytic approaches to identify reliability related characteristics of enunciations, Section \ref{sec:compapproach} presents the computational approach followed, and Section \ref{sec:conclusions} summarises the conclusions and proposes a roadmap for future work.

\section{Related Work}
Relevant, yet different, approaches of combining psychological and social dimensions with computational methods include computational psycholinguistics, personality traits, behavioural analysis, emotional states and cognitive psychology methods. Compared to all those approaches, our approach is fundamentally different since we adopt a psychoanalytic perspective; in particular, we employ the powerful notion of Lacanian discourse types. To the best of our knowledge, this is the first attempt of systematically bringing together psychoanalysis and computing. We believe that such a psychoanalytic approach is eventually more effective compared to the previously mentioned methods, since it addresses deeper, fundamental elements of human personality, behaviour and expression which usually escape methods operating at a ``higher'' conscious layer.  
  
Having stressed this general novelty of our research methodology, we below discuss relevant, recent research related to the particular case study (fake news detection) which we use in order to exemplify our method.

A fundamental approach of combining psychology with computational linguistics (based on abstract formulations of phrases via a collection of finite-state transition networks) is described in \cite{Kaplan}; in particular, the author envisions sophisticated natural language technologies as a key factor for improving the (rather poor) performance of current conversational systems used by modern technology. The abundant availability of massive data along with effective AI methods (including deep learning) is expected to further facilitate this vision. We note that, although the notion of conversation is directly relevant to the notion of discourse, the approach taken in that paper is more limited (psychological) than the directly psychoanalytic attempt we pursue in our research. 

For the more concrete aspect of detecting misinformation in online social networks, \cite{Kumar} suggests the application of cognitive psychology concepts. An efficient algorithm for detection of spread of misinformation in Twitter is proposed, based on text and network-wide qualities such as the consistency of message, the coherency of message, the credibility of source and the general acceptability of message in the network. Again, no psychoanalytic elements are considered when evaluating the qualitative properties of the text. Also, the use of objective, global information is employed, in contrast to our approach which focuses on each text separately (however, our method can also be extended to include global information about texts). 
  
Psychological factors (in particular, emotions as expressed in Reddit conversations) are addressed in \cite{Guo}, to propose a model for passively detecting mental disorders. The suggested model is based entirely on emotional states and the transitions between these states identified in Reddit posts, in contrast to content-based representations (e.g., n-grams, language model embeddings etc.) in the relevant state of the art. The scope is to overcome the domain and topic bias of content-based representations, towards more general applicability. Our approach aims to avoid a content-specific bias, focusing on underlying qualities of texts captured by the discourse type identification. In fact, discourse types are an even more ``primitive'' aspect of texts, more so than emotions; so the generality of our method might be broader. 

In another psychology-based research for fake news detection, a behavioural analysis approach is taken \cite{Cardaioli}. In particular, the authors use supervised learning algorithms to profile fake-news spreaders, based on the combination of Big Five personality traits and stylometric features. The method is evaluated on a tweeter dataset in both English and Spanish languages. In a similar spirit, \cite{Sampat} aims to understand the motivations for sharing fake news and the corresponding personality traits among social media users in India; in particular, the findings suggest that the passing time, information sharing and socialisation gratifications lead to instant sharing news. Also, people who exhibit extraversion, neuroticism and openness share news on social media platforms instantly; in contrast, agreeableness and conscientiousness personality traits lead to authenticating news before sharing. We note that that work focuses on fake news spreading, not fake news identification itself. Also, as stated previously, although those works exhibit a psychology-based flavour too, our own research is based on psychoanalytic methods, particularly for identifying the power of characteristic Lacanian discourse types to identify specific characteristics in texts in interactive media as, for example, the fake news identification that we selected to develop the case study of this work. 
\label{sec:relatedwork}

\section{Psychological and Psychoanalytic Approaches Taken}
\label{sec:psyapproaches}
In this section, we describe two distinct approaches to potentially infer reliability-related information from user data. The first one is based on the psychological concept of personality traits, and the second one on the psychoanalytical concept of Lacanian discourses.
\subsection{Personality Type Prediction}
\label{sec:personalityapproach}
Two potential answers to the question "What makes people unique and different from one another?" are motives or traits \cite{Oliver17}. Regarding motives, we can find the work "Motivation and personality: Handbook of thematic content analysis" \cite{smith_1992_14} and Murray's list of needs \cite{Murray2007_15}. However, no computational approach has been implemented or proposed. Personality traits and types, on the other hand, have numerous theories, taxonomies, and frameworks available, of which the most well known are the Big Five \cite{Rothmann2003_16} and the Myers–Briggs-Type-Indicator (MBTI) \cite{Boyle1995_5}. More specifically, the trait approach, rather than the motives approach, is still the most widely used and most accepted conceptual framework for describing human personality and successfully predicting human behavior, and has been implemented and applied multiple times in computer-based applications, such as in social media advertising \cite{Clark2014_2}, trait prediction from facial images \cite{Kachur2020_3}, and many others. Both these systems are intended to indicate or predict the subject's personality, and thus his or her behavior and preferences, by breaking them down into many dimensions, or groups. In this research, we are focusing mainly on the traits approach and particularly the MBTI system.

\begin{figure}[!htb] 
\begin{center}
 \includegraphics[keepaspectratio=true, width=0.4\textwidth]{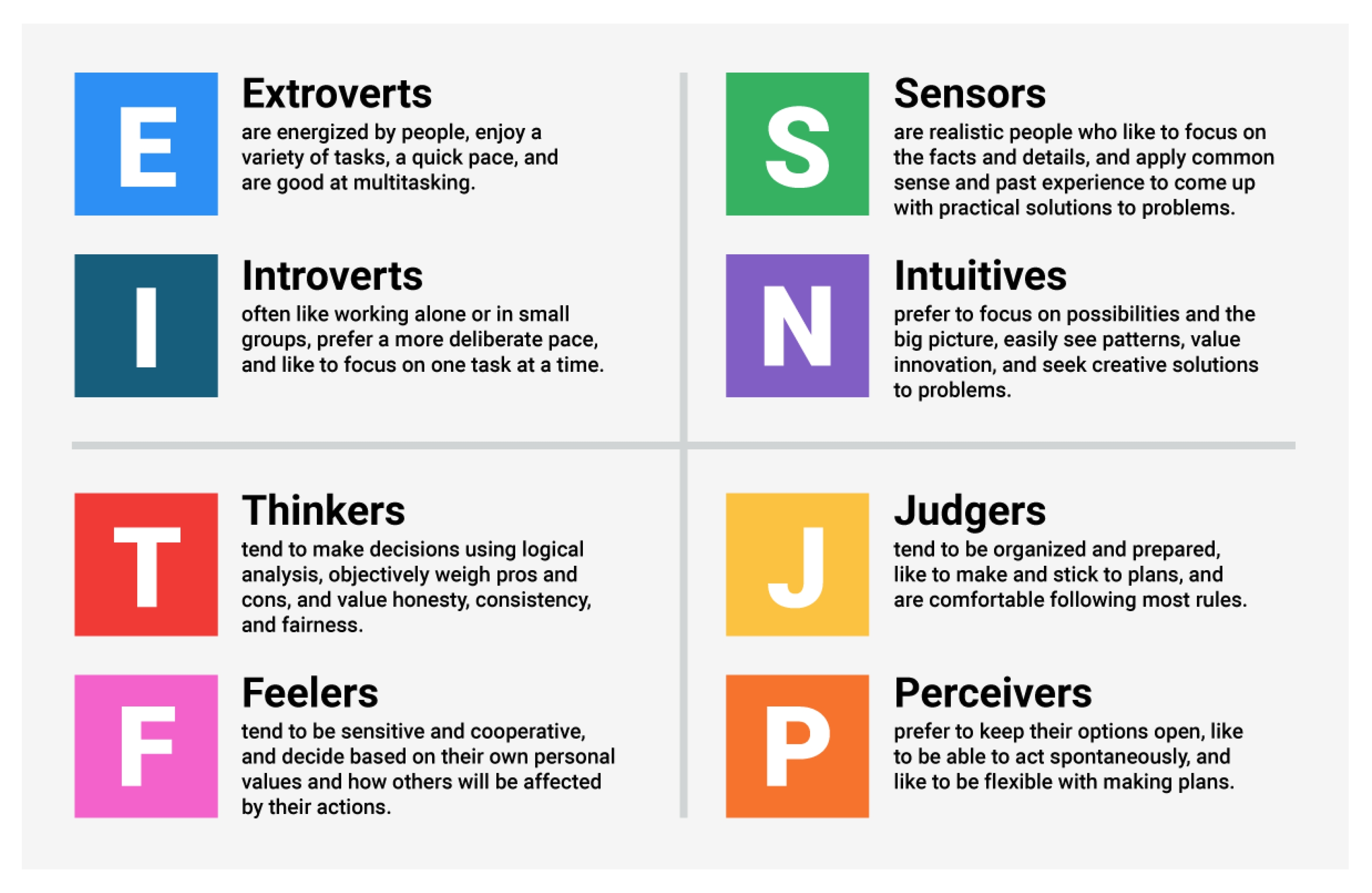} 
 \end{center}
\caption{ Personality types groups}
\label{fig:mbti_1}
 \end{figure}

The Big Five personality traits comprise the following: extraversion, conscientiousness, agreeableness, openness to new experiences, and neuroticism \cite{Rothmann2003_16
}. Meanwhile, the MBTI personality model describes an individual's preferences/behaviour in four dimensions/groups which are also illustrated in Figure \ref{fig:mbti_1}: 
\begin{itemize}
    \item Extraversion-Introversion: This dimension measures how a person gets their energy. Extraverts are energized by being around people, while introverts are energized by being alone.
    \item Sensing - Intuition: This dimension measures how a person takes in information. Sensors rely on their five senses to gather information, while Intuitives rely on their gut feelings and intuition
    \item Thinking- Feeling: This dimension measures how a person makes decisions. Thinkers use logic and reason to make decisions, while Feelers use their emotions and values.
    \item Judging - Perceiving: This dimension measures how a person lives their life. Judgers like structure and order, while Perceivers are more flexible and spontaneous.
\end{itemize}

This way, each individual is classified in terms of one of 16 possible four-letter codes, like ESFJ, indicating a person who may:
\begin{itemize}
    \item  Extraverted (E): is more concerned with the outside world of people and things than with the inner world of ideas
    \item Sensing (S): would rather work with known facts and solid experience than explore possibilities or meanings
    \item Feeling (F): steps into situations to weigh human values and motives. Prefers to make decisions on the basis of values. 
    \item Judging (J): prefers a planned, determined orderly way of life to a flexible spontaneous.

\end{itemize}

MBTI was developed by Isabel Briggs Myers in the 1940s \cite{Boyle1995_5} to implement Jung’s \cite{9780691097701_4} theory of psychological types and the results from such an approach are obtained classically from test booklets, answer sheets, and score keys, and are produced professionally \cite{Cattell1943_1} \cite{Boyle1995_5}. In Sections \ref{sec:preliminaryevaluation} and \ref{sec:compapproach} we discuss and use some computational ways of determining personality types using semantic and linguistic analysis and present the results from a Fake/Real news detection application. 

\subsection{Lacanian Discourses}
\label{sec:lacanianapproach}
\makeatletter
\newcommand{\superimpose}[2]{%
  {\ooalign{$#1\@firstoftwo#2$\cr\hfil$#1\@secondoftwo#2$\hfil\cr}}}
\makeatother

\newcommand{\veewedge}{\mathpalette\superimpose{{\vee}{\wedge}}}
\newcommand{\lessgreater}{\mathpalette\superimpose{{<}{>}}}
\newcommand{\strikeQ}{\mathpalette\superimpose{{\text{---}}{Q}}}

\newcommand{\Xarrows}{\mathpalette\superimpose{{\nearrow}{\nwarrow}}}

\newcommand{\matheme}[4]{\left\uparrow\dfrac{#1}{#2} \hspace{.3cm}\overrightarrow{\Xarrows}\hspace{.3cm} \dfrac{#3}{#4}\right\downarrow}

\newcommand{\master}{\matheme{\text{S1}}{\$}{\text{S2}}{a}}

\newcommand{\university}{\matheme{\text{S2}}{S1}{\text{a}}{\$}}

\newcommand{\analyst}{\matheme{\text{a}}{S2}{\text{\$}}{S1}}

\newcommand{\hysteric}{\matheme{\text{\$}}{a}{\text{S1}}{S2}}

\newcommand{\discourses}{\matheme{\text{Agent}}{\text{Truth}}{\text{other}}{\text{Production}}}

The Four Discourses theory constitutes an attempt of formalisation of the different ways people relate to each other and the economy of knowledge and enjoyment in social relationships. The Lacanian framework defines a more complex representation of the roles assumed by two interacting parties, formulating four discrete discourse types \cite{Lacan20}, \cite{Lacan22} and \cite{bailly}: {\bf Discourse of the Master} -- Struggle for mastery/domination/penetration; {\bf Discourse of the University} -- Provision and worship of ``objective'' knowledge -- usually in the unacknowledged service of some external master discourse; {\bf Discourse of the Hysteric} -- Symptoms embodying and revealing resistance to the prevailing master discourse; {\bf Discourse of the Analyst } -- Deliberate subversion of the prevailing master discourse.

Later, Lacan defined an additional fifth discourse, which is not  considered in this work, the discourse of the Capitalist \cite{contri2008}, where the subject is commanded to enjoy in the form of commodities. 

\begin{figure}[!htb] 
\begin{center}
 \includegraphics[keepaspectratio=true, width=0.4\textwidth]{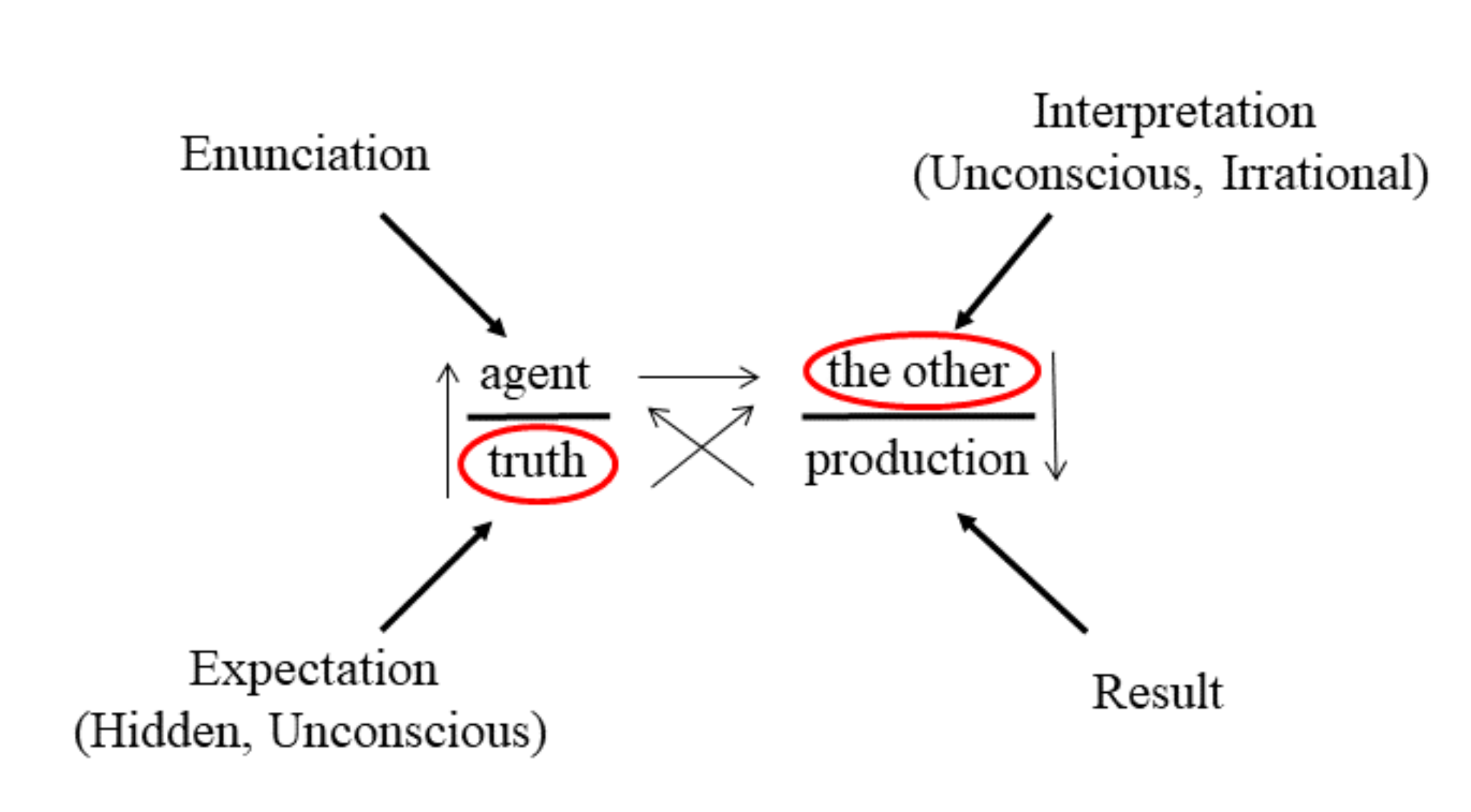} 
 \end{center}
 \caption{General representation of the Lacanian discourses.}
 \label{fig: fourdiscourses}
 \end{figure}

Figure \ref{fig: fourdiscourses} presents the general representation of the Lacanian discourses. The four places of the discourse: the ``agent'', the giver of the discourse; the ``other'', the one to whom the discourse is addressed; under the message of the agent is hidden the ``truth'', which is masked by the official statement; and hidden under the other is the ``production'', or what the agent gets out of the relationship. It is important to realise that the influence of the hidden ``truth'' upon the ``agent'' and the enunciation interpretation performed by the ``other'' are subjective unconscious processes of the communicating parties

It is possible to draw a parallel between the terms of a discourse and the components of a communication process, in such a way that the dynamics of a given discourse, i.e., the internal relations between elements arranged in different places, can serve to characterise the dynamics of a given media process. 

\FloatBarrier

The elements of a discourse or the roles assumed by the communicating parties are the following:
 \begin{itemize}
 	\item {\bf S1}: the master signifier represents the true essence of the subject. It may be summarised by \emph{who I am}.
	\item {\bf S2}: represents the knowledge of the subject. It may be summarised by \emph{what I know}.
	\item {\bf a}: represents the object cause of desire. It may be summarised by \emph{what I want}.
	\item {\bf \$}: represents the barred subject castrated by the language. It may be summarised by \emph{what I speak}.
 \end{itemize}

The discourses are defined by the position each element occupies in the general representation depicted in Figure \ref{fig: fourdiscourses}. Each discourse representation is called, in the Lacanian jargon, a \emph{matheme}. The definitions given below are reproduced from \cite{bailly}.

\begin{quote}
$\master$ {\bf Master Discourse:} the master signifier ({\bf S1}) (of the Master) is the agent of communication and instead of addressing the other ({\bf a}), addresses the ``knowledge'' ({\bf S2}) in his/her place; in other words, the Master is addressing the other not as a Subject but in his/her functional role, because of his/her ability or knowledge (S2). The truth of the master signifier is actually a barred Subject ({\bf \$}) and, as lacking as everyone else, is hidden; however, beneath that master signifier ({\bf S1}), the barred Subject ({\bf \$}) is enjoying the production of the knowledge (what comes from {\bf a}).
\end{quote}

\begin{quote}
$\university$ {\bf University Discourse:} the discourse of the University makes a point about the functioning of institutions, and by extension of the individuals within them in their capacity of incarnating the institution. Knowledge ({\bf S2}) occupies the place of the agent, which addresses itself to the object cause of desire ({\bf a}), as the desire for knowledge is the supposed reason why the student is there. However, in this relationship, one can see that the \emph{objet petit} {\bf a} is also, and perhaps as importantly, fed into by the master signifiers of the institution ({\bf S1}), and these contribute endlessly to the castratedness of the Subject of the student. In addressing the knowledge not to the Subject, but to the object cause of desire of the Subject, what is ``produced'' in, in fact, more castratedness ({\bf \$}). Beneath the appearance of dispensing knowledge ({\bf S2}), the University controls the Subject ({\bf \$}) by means of its master signifiers ({\bf S1}) and enjoys the ``production'' of the castrated student ({\bf \$}). The institution is also guilty of giving the impression to the student that by careful attention and absorption of its master signifiers, she/he may overcome his/her castration. This is a system of functioning that is common to all institutions: corporations, professions, and government departments, indeed in any institution where ``knowledge'' ({\bf S2}) in some form takes the place of the agent which addresses discourse, and acts as a lure to the others desire ({\bf a}).
\end{quote}

\begin{quote}
$\analyst$ {\bf Analyst Discourse:} in the discourse of the Analyst, the Analyst has accepted becoming symbolically, the \emph{objet petit} {\bf a} of the analysand. This is one of the most usual roles the analyst has to accept; he/she is an empty mirror upon which everything may be reflected, and when in full transference, the analysand ({\bf S1}) will be addressing his/her object cause of desire ({\bf a}). In this case, in his/her role as the \emph{objet petit} {\bf a}, the Analyst addresses his/her discourse to the castratedness ({\bf \$}), the anxiety, of the patient, and his/her questioning pushes the analysand to produce a master signifier ({\bf S1}) which is reflected back to the Analyst, while the hidden knowledge of the Analyst ({\bf S2}), in the place of truth, is fed into the castrated Subject ({\bf \$}), fuelling the production by it of master signifiers ({\bf S1}). The analysand will discover that the knowledge of its own desire ({\bf S2} hidden beneath {\bf a}) is not held by the Analyst but revealed through its master signifier ({\bf S1}). The Analyst does not adopt a position of power like the Master, or of knowledge, like the University, and because of that, is often considered subversive by institutions.
\end{quote}

\begin{quote}
$\hysteric$ {\bf Hysteric Discourse:} one doesn't have to be hysterical in the clinical sense to hold the Discourse of the Hysteric; indeed, Lacan made it clear that this type of discourse in non-hysterical people, is precisely what leads to true learning. The agent of the discourse is the castrated ({\bf \$}) shortage of the Hysteric; hidden beneath its bar is his/her object cause of desire ({\bf a}). This barred Subject ({\bf \$}), driven by its \emph{objet petit} {\bf a}, addresses the master signifiers of the other ({\bf S1}), which respond with the production of knowledge ({\bf S2}), beneath the bar. It is to the master signifier ({\bf S1}) that the Hysteric ({\bf \$}) addresses his/her questions, but she/he receives as an answer only the knowledge ({\bf S2}) of that person, which the Hysteric enjoys for want of anything better, although these answers never constitute a satisfactory response to his/her desire ({\bf a}). The Discourse of the Hysteric is held by anyone who is on the path to knowledge. It is a position that requires perfect acceptance of one's ignorance, no great desire to pretend to any other status, and a hunger for the object cause of desire\footnote{It is a characteristic of the Hysteric discourse to be charged with emotion.}
\end{quote}

Precisely, the idea is that {\bf by distinguishing the discourse type, the $\ll$ truth $\gg$ status of an enunciation can be qualified}; i.e., it is the formal characteristics of the discourse which will inform if the truth is based on authority, on documented sources, on the needs to identify the essence of the other, or on the needs to dig for information, respectively.  {\bf The formal characteristics will be a key to measure the ``reliability'' likelihood of the enunciation}. While the personality traits tell us about the speaker, the Lacanian discourses tell us about what is said. A trained psychoanalyst can detect the above signifiers based on semantical analysis of the language.

It can be argued how the Lacanian discourse approach can be extended to groups with more than two interacting parties. The answer to this concern comes from recognising that psychoanalysis is a general framework for the interpretation of situations expressed in any format and by any number of people. Since its development by Freud it has been stated and shown that it can be used to interpret works of art \cite{freud1914}, to analyse social situations \cite{freud1921}, \cite{freud1933} or to conjecture about the future of civilisation \cite{freud1927}, \cite{freud1930}. These are just a few examples of Freud's works applying Psychoanalysis techniques out of the psychoanalytical setting of a patient and an analyst. 

Lacanian discourses are a formal framework to apply psychoanalytical concepts to interpret any real life situation. As the aforementioned Freud's works, they can be used as a reference model to interpret works of art, social situations, and so on.

Nevertheless, the association of an interaction with one of the discourses is not an easy task and has to be dealt with caution. An example that illustrates how a single media phenomenon can be seen from various angles according to the discourse theory is provided by \cite{Castro}: ``When Google scans the Internet collecting information from each site, we are in the discourse of University. When it meets our demand for providing results, we are in the discourse of hysteria. When we deify it, we are in the discourse of the master. When it computes our data and customises the results it offers us, as if it knew us, knew our preferences and anticipate what we want, we are in the discourse of capitalism''. Therefore, it is paramount that the context is well defined before proceeding to association, so that the stakeholders of the discourse are clearly identified. 

Context depends on the kind of application. The knowledge of the application implies the knowledge of the context and establishes the contextual elements to be considered.

\section{Preliminary Evaluation of the Psychological and Psychoanalytic Approaches}
\label{sec:preliminaryevaluation}
In this section, we empirically evaluate the potential of the Psychological and Psychoanalytic approaches described in Sections \ref{sec:personalityapproach} and \ref{sec:lacanianapproach} to identify reliability related characteristics of enunciations. 

\subsection{Dataset used for evaluation}

The selected dataset was obtained from Kaggle \footnote{\url{https://kaggle.com/datasets/rchitic17/real-or-fake, Accessed May. 25, 2022}} and consists of a number of news headlines and content that have been used to develop Real-Fake news algorithms as, for example by \cite{koury2019}; they obtained a 97\% accuracy of Fake/Real news prediction utilizing the headlines and content, using the  state-of-the-art NLP methods. In our approach, from this dataset, only the headlines were used and  submitted to the personality traits assignment algorithm developed by \footnote{\url{https://github.com/wiredtoserve/datascience/tree/master/PersonalityDetection, Accessed May. 25, 2022}}. Each headline also has an associated label. When the label is equal to 0, the corresponding news is Real, otherwise it is Fake. 

\begin{figure}[!htb] 
\begin{center}
\includegraphics[keepaspectratio=true, width=0.4\textwidth]{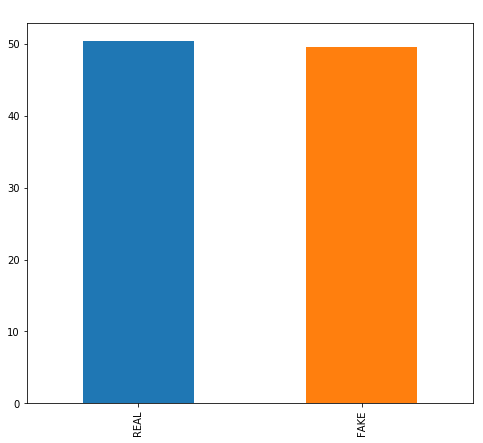} 
\end{center}
\caption{ Label Real=0 and Fake=1 distribution}
\label{fig:lbl_distro}
\end{figure}

\FloatBarrier

For the needs of this application after analyzing the content we decided to remove any headline which was less than 4 words, as most of them don't give any semantic information if it is fake news or not. The final dataset, containing 5860 unique \emph{headlines}, is balanced, ideal for the task of applying the MBTI theory and the labels are equally distributed as shown in Figure \ref{fig:lbl_distro}. Finally, for illustration purposes, Table \ref{tab:ds_sample} contains three  random selected rows from the dataset:

\begin{table}[!htb] 
\caption{Three random selected headlines from the dataset for illustration purposes}
\label{tab:ds_sample}
\begin{tabular}{||p{.385\textwidth} c||}
 \hline
 {\bf Headline} & {\bf Label} \\ [0.5ex] 
 \hline
 Police Turn In Badges Rather Than Incite Violence Against Standing Rock Protestors & 1 \\ 
 \hline
 Republican race enters a new volatile phase & 0 \\
 \hline
  First Take Wall Street bids goodbye to June hike
 & 0 \\
 \hline
\end{tabular}
\end{table}

\FloatBarrier



\subsection{Evaluation of the personality traits (MTBI) approach}
\label{sec:preliminary-mtbi-evaluation}

An algorithm was developed in the R language to evaluate the potential of the personality traits approach to identify reliability related characteristics of the dataset's \emph{headlines}. R is a free software environment for statistical computing and graphics. It compiles and runs on a wide variety of UNIX platforms, Windows and MacOS\footnote{R can be downloaded from \url{https://www.r-project.org}.}. The main steps of the algorithm are:
\begin{itemize}
	\item[Step 1:] the dataset is read and stored in a dataframe called {\tt raw.comments.df}. 
	
	\item[Step 2:] the dataset is split into test and evaluation datasets, {\tt test.comments.df} and {\tt eval.comments.df}. The test dataset comprises 40\% of the \emph{headlines}.
	
	The personality traits are designated as:
	\begin{itemize}
		\item[-] Extroverts - Introverts -- EI;
		\item[-] Sensors - Intuitives -- SN;
		\item[-] Thinkers - Feelers -- TF;
		\item[-] Judgers - Perceivers -- JP.
	\end{itemize}
	
	\item[Step 3:] the test dataset is split into two datasets, one containing the Real \emph{headlines} ({\tt test.comments.0.df}), i.e, the \emph{headlines} for which Label = 0, and one containing the Fake \emph{headlines}({\tt test.comments.1.df}), i.e., those headlines for which Label = 1.
	
	The number of Real and Fake test \emph{headlines} are 1156 and 1188, respectively. 
	
	\item[Step 4:] the cumulative distribution function (cdf) for each personality trait given the value of the label,  given by Eq. \ref{eq:cdf1}, is evaluated.
\end{itemize}

	\begin{equation} \label{eq:cdf1}
	\text{cdf}_{x}(a | v) = \text{Prob}(x \leq a | \text{label} = v)
	\end{equation}
	where $x \in\{EI, SN, TF, JP\}$, $0 \leq a \leq 1$ and $v \in\{0, 1\}$.

\begin{itemize}
	\item[Step 5:] using the Bayes theorem, the probability of the label assuming a determined value given the probability of a personality trait is less than or equal to a certain value  is evaluated by Eq. \ref{eq:plabel}.
\end{itemize}

\begin{equation} \label{eq:plabel}
	\text{Prob}(\text{label} = v | x \leq a) = \dfrac{\text{Prob}(x \leq a | \text{label} = v)}{\sum_{v \in\{0, 1\}}\text{Prob}(x \leq a | \text{label} = v)}
	\end{equation}
	where $x \in\{EI, SN, TF, JP\}$, $0 \leq a \leq 1$ and $v \in\{0, 1\}$.
	
\begin{figure}[!htb] 
\begin{center}
 \includegraphics[keepaspectratio=true, width=0.4\textwidth]{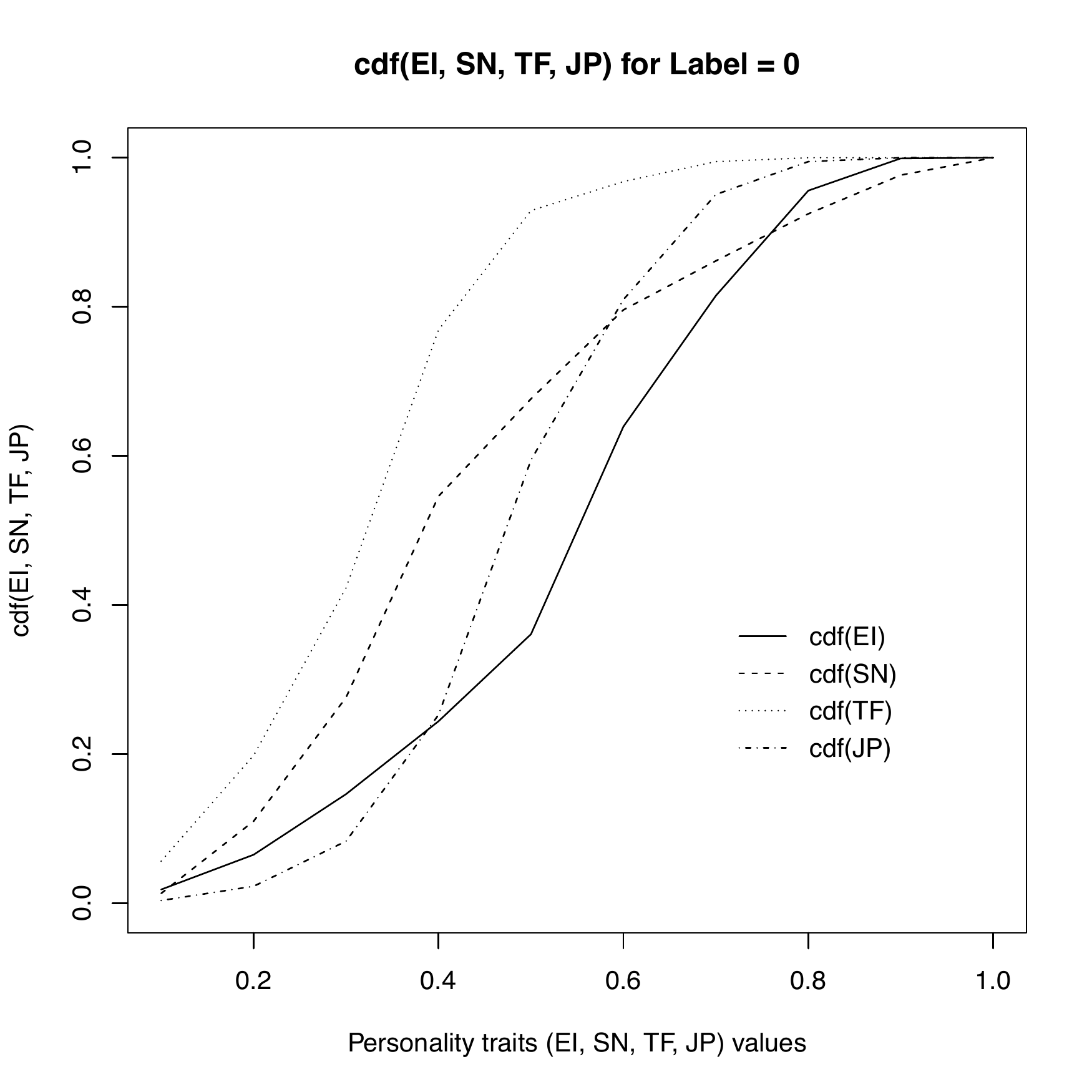} 
 \end{center}
 \caption{Cumulative distribution function for each personality trait and Label = 0.}
 \label{fig:cdf0}
 \end{figure}

\begin{figure}[!htb] 
\begin{center}
 \includegraphics[keepaspectratio=true, width=0.4\textwidth]{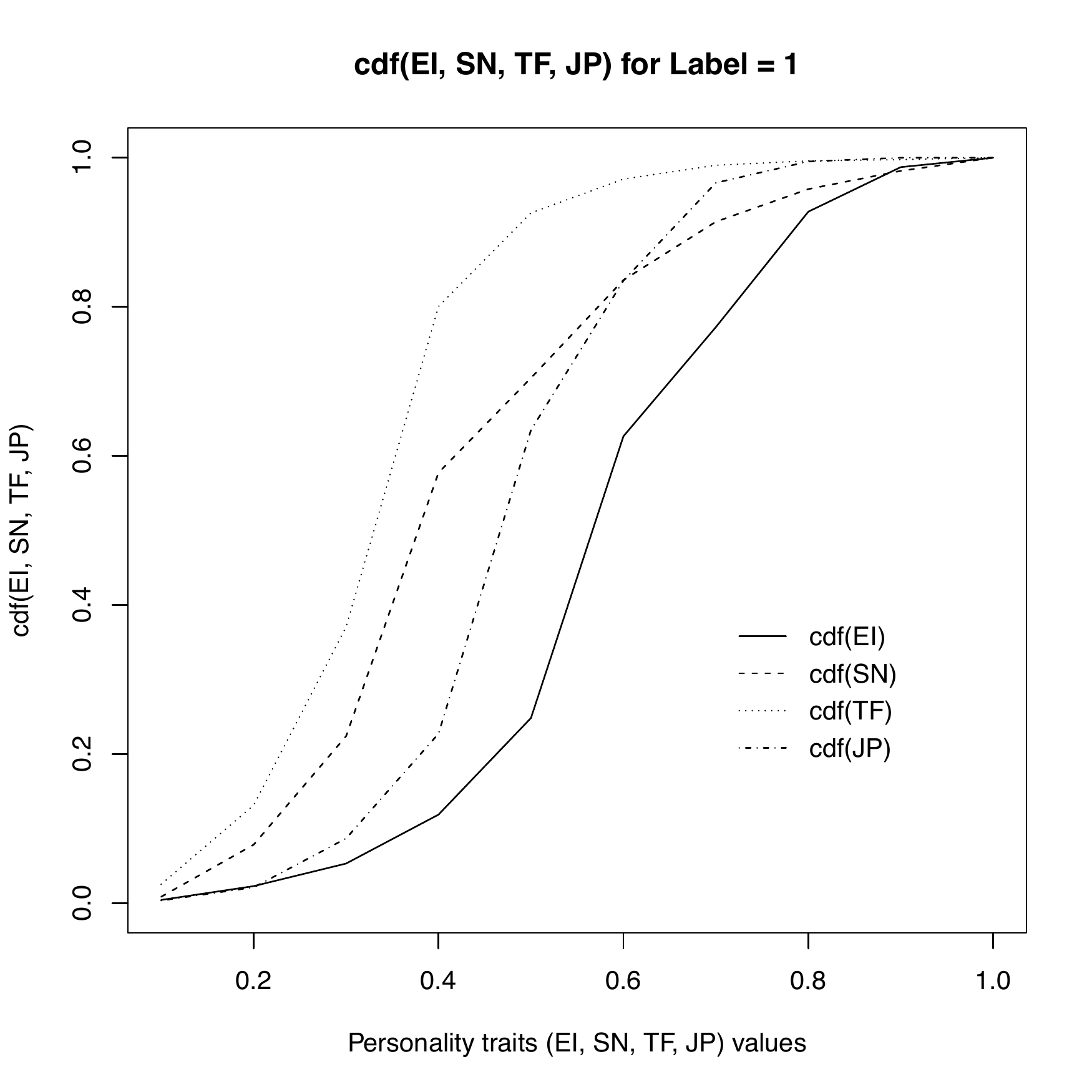} 
 \end{center}
 \caption{Cumulative distribution function for each personality trait and Label = 1.}
 \label{fig:cdf1}
 \end{figure}

Figures \ref{fig:cdf0} and \ref{fig:cdf1} show the cumulative distribution function for each personality trait, Label = 0 and Label =1. It can be seen that the distributions are very similar indicating that it may not be possible to differentiate between the labels using the personality traits.

\FloatBarrier

\FloatBarrier

\begin{figure}[!htb] 
\begin{center}
 \includegraphics[keepaspectratio=true, width=0.45\textwidth]{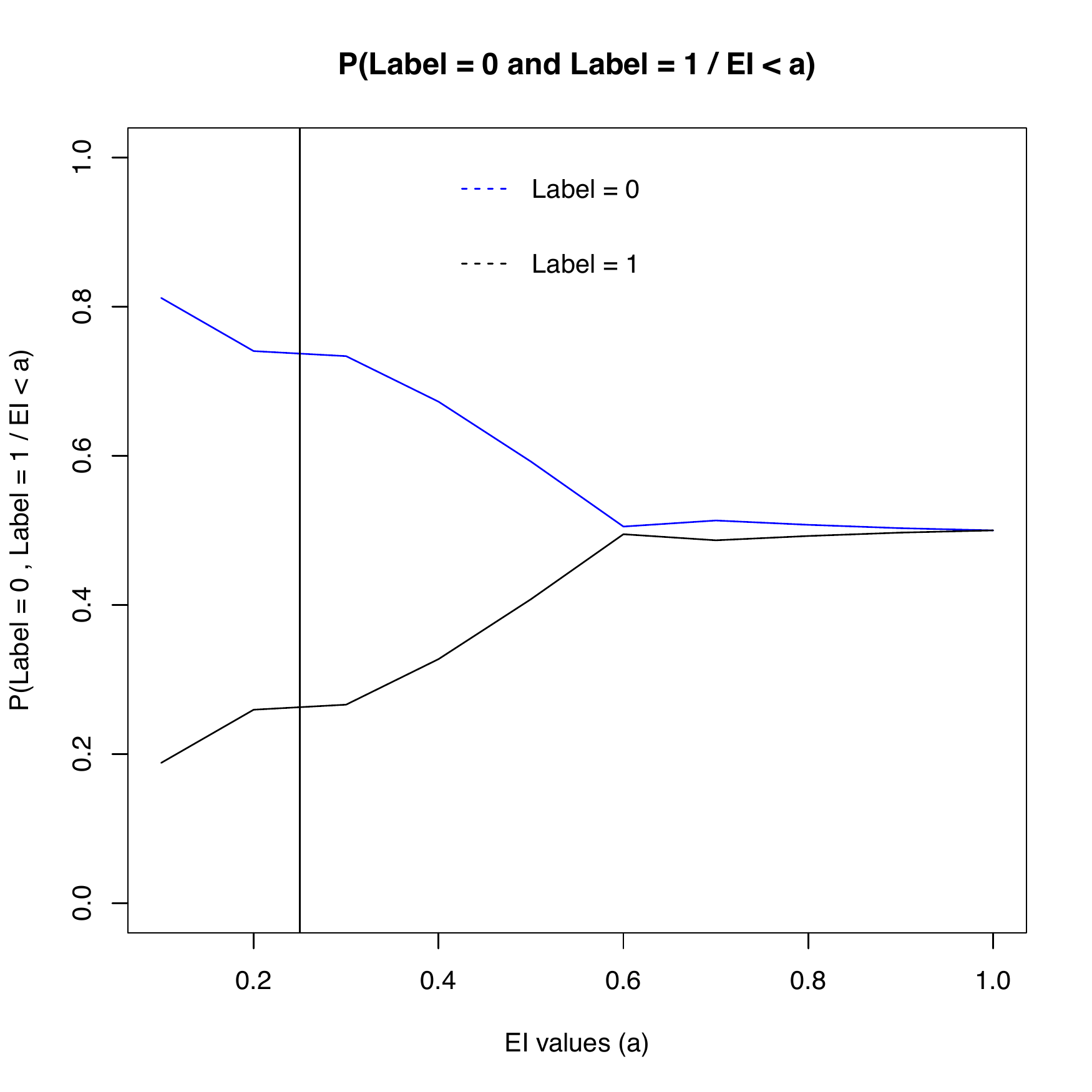} 
 \end{center}
 \caption{Probability of the label assuming a determined value given the probability of a personality trait is less than or equal to a certain value.}
 \label{fig:prob-label-EI}
 \end{figure}

\begin{figure}[!htb] 
\begin{center}
 \includegraphics[keepaspectratio=true, width=0.45\textwidth]{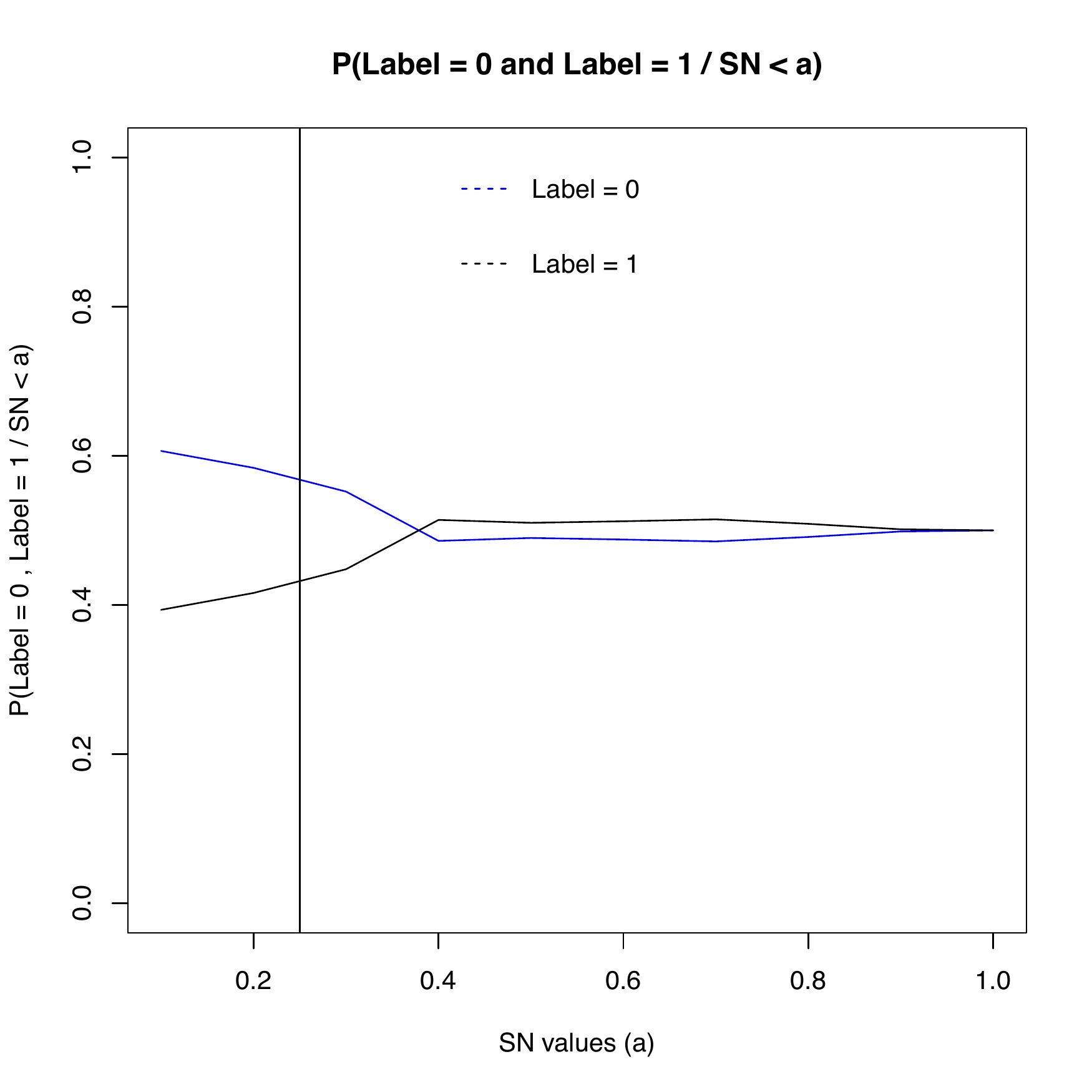} 
 \end{center}
 \caption{Probability of the label assuming a determined value given the probability of a personality trait is less than or equal to a certain value.}
 \label{fig:prob-label-SN}
 \end{figure}

\begin{figure}[!htb] 
\begin{center}
 \includegraphics[keepaspectratio=true, width=0.45\textwidth]{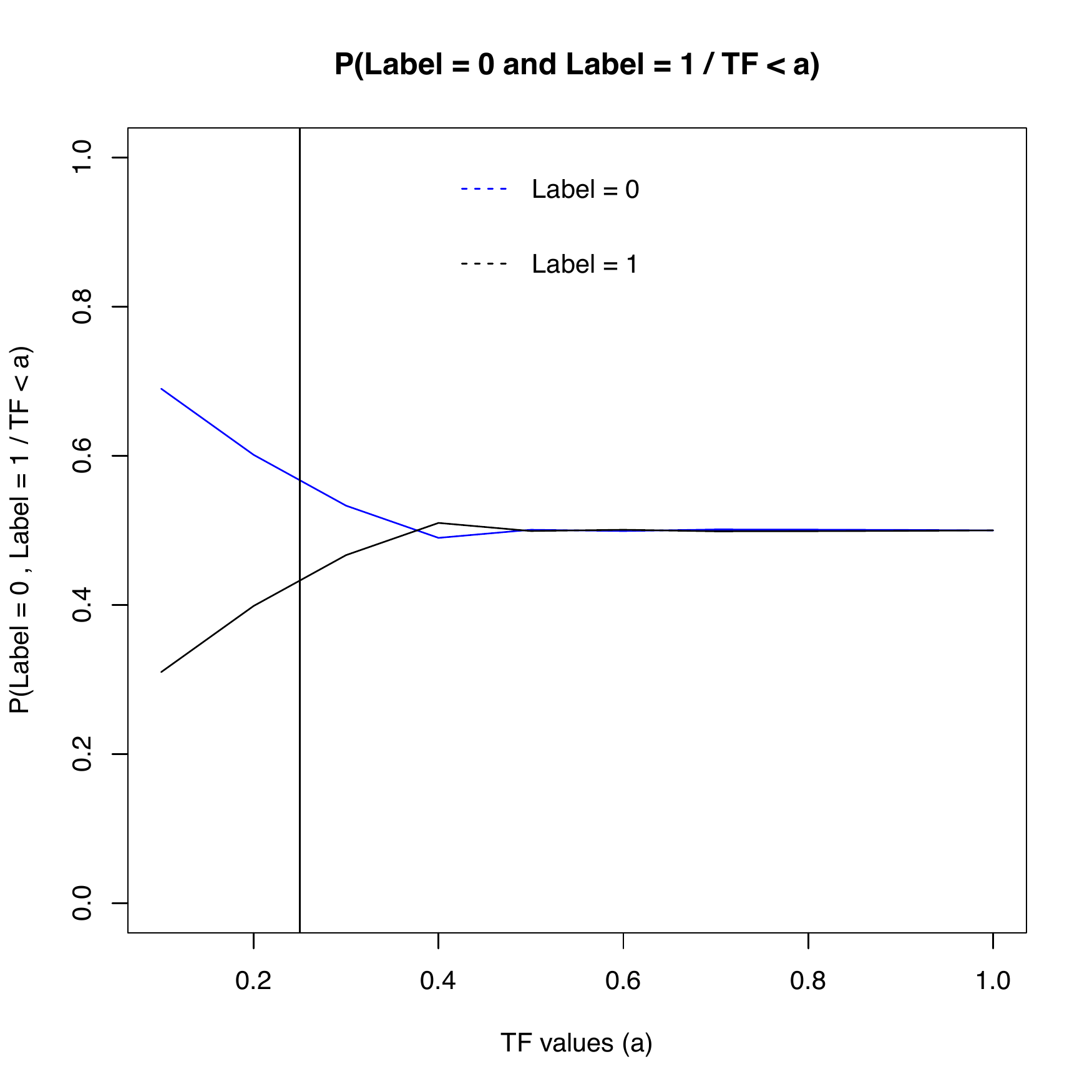} 
 \end{center}
 \caption{Probability of the label assuming a determined value given the probability of a personality trait is less than or equal to a certain value.}
 \label{fig:prob-label-TF}
 \end{figure}

\begin{figure}[!htb] 
\begin{center}
 \includegraphics[keepaspectratio=true, width=0.45\textwidth]{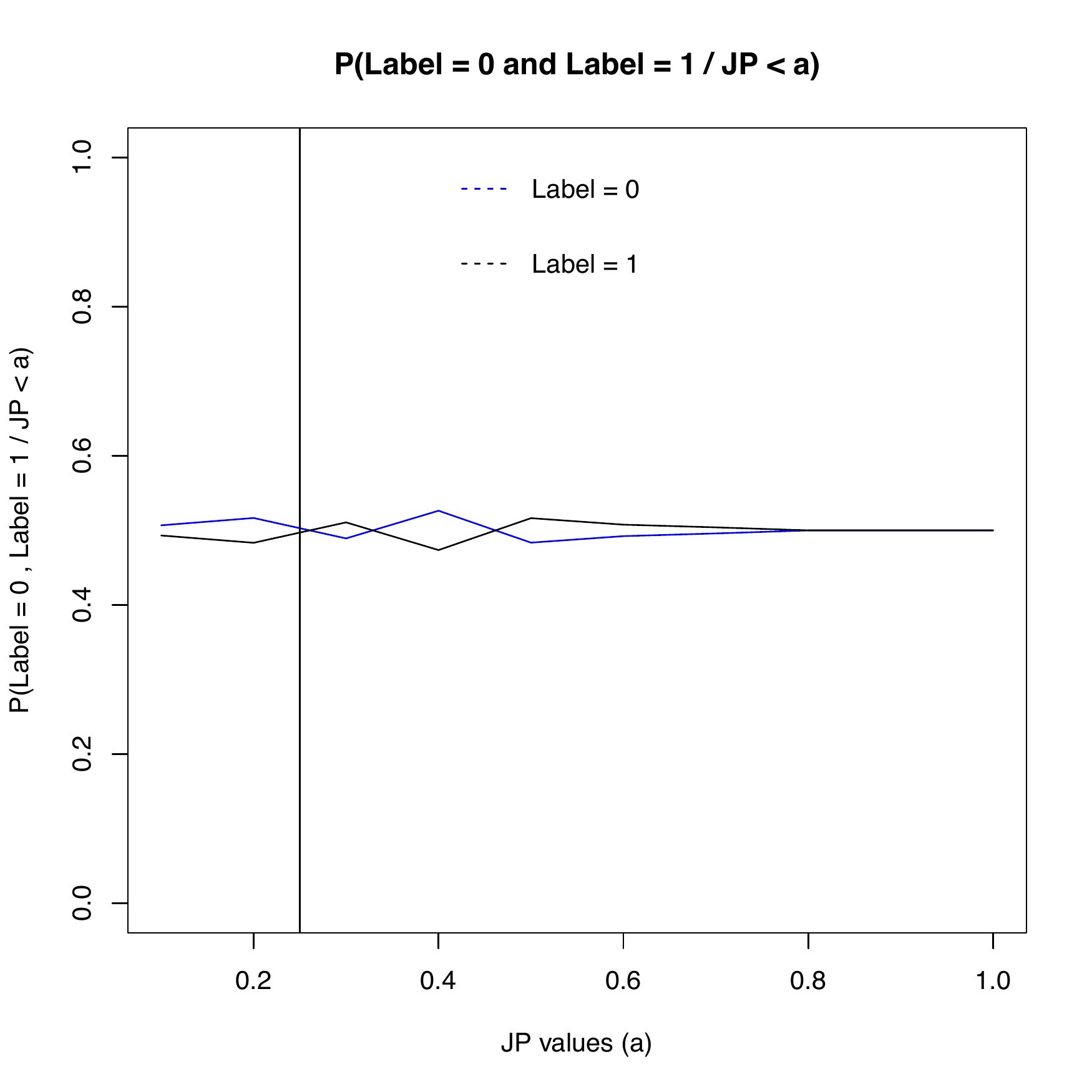} 
 \end{center}
 \caption{Probability of the label assuming a determined value given the probability of a personality trait is less than or equal to a certain value.}
 \label{fig:prob-label-JP}
 \end{figure}

Figures \ref{fig:prob-label-EI}, \ref{fig:prob-label-SN}, \ref{fig:prob-label-TF} and \ref{fig:prob-label-JP} show the probability of the label assuming a determined value given the probability of a personality trait (EI, SN, TF or JP) is less than or equal to a certain value.

\FloatBarrier

\FloatBarrier

\FloatBarrier

\FloatBarrier



It can be seen in Figures \ref{fig:prob-label-EI}, \ref{fig:prob-label-SN}, \ref{fig:prob-label-TF} and \ref{fig:prob-label-JP}  that: i) JP does not help at all to infer the labels values; ii) from a certain value of the EI, SN and TF personality trait it is not possible to infer the labels values. It could conjectured that low values personality traits have some differentiation power.

\begin{itemize}
	\item[Step 6:] a threshold limit of 0.25 was defined for the personality traits values and several simple algorithms were developed to evaluate their differentiation power. The first one is based on the total number of low value personality traits, assuming at least two traits with low values. The following three algorithms, each focuses on a single trait of low value. The final one requires that one of the traits has low value.
\end{itemize}

The obtained results are summarised as follows:

\paragraph*{{\bf Global evaluation:}} total number of evaluation \emph{headlines}: 3516.

\begin{itemize}
	\item[-]tested condition: if the total number of low value personality traits is greater than {\bf 1} then the predicted label is set to 0; otherwise the predicted label is set to 1;
	\item[-]total number of errors: 1674;
	\item[-]accuracy: 52.39\%
\end{itemize}

\begin{itemize}
	\item[-]tested condition: if the total number of low value personality traits is greater than {\bf 2} then the predicted label is set to 0; otherwise the predicted label is set to 1;
	\item[-]total number of errors: 1792;
	\item[-]accuracy: 49.03\%
\end{itemize}

\begin{itemize}
	\item[-]tested condition: if the total number of low value personality traits is greater than {\bf 3} then the predicted label is set to 0; otherwise the predicted label is set to 1;
	\item[-]total number of errors: 1796;
	\item[-]accuracy: 48.93\%
\end{itemize}

\paragraph*{{\bf Low EI values performance:}} number of \emph{headlines} with low EI personality trait values: 55.

\begin{itemize}
	\item[-]tested condition: if the \emph{headline's} EI  personality trait value is less than the threshold  limit and the SN and TF values are greater than the threshold limits then the predicted label is set to 0; otherwise the predicted label is set to 1;
	\item[-]number of correct predictions: 37;
	\item[-]accuracy: 62.27\%
\end{itemize}

\paragraph*{{\bf Low SN values performance:}} number of \emph{headlines} with low SN personality trait values: 364.

\begin{itemize}
	\item[-]tested condition: if the \emph{headline's} SN  personality trait value is less than the threshold  limit and the EI and TF values are greater than the threshold limits then the predicted label is set to 0; otherwise the predicted label is set to 1;
	\item[-]number of correct predictions: 210;
	\item[-]accuracy: 57.69\%
\end{itemize}

\paragraph*{{\bf Low TF values performance:}} number of \emph{headlines} with low SN personality trait values: 582.

\begin{itemize}
	\item[-]tested condition: if the \emph{headline's} SN  personality trait value is less than the threshold  limit and the EI and TF values are greater than the threshold limits then the predicted label is set to 0; otherwise the predicted label is set to 1;
	\item[-]number of correct predictions: 299;
	\item[-]accuracy: 51.37\%
\end{itemize}

\paragraph*{{\bf Low personality traits values global performance:}} total number of \emph{headlines} with low personality trait values: 1001.

\begin{itemize}
	\item[-]tested condition: if any of the last three tested conditions is satisfied then the predicted label is set to 0; otherwise the predicted label is set to 1;
	\item[-]number of correct predictions: 546;
	\item[-]accuracy: 54.54\%
\end{itemize}

The results suggest that the use of personality traits to predict if a \emph{headline} is Real or Fake leads to no better than a random assignment.  Even the 62.27\% accuracy obtained for the low EI values case cannot be considered a good result as just 55 \emph{headlines} fall into this category. 

As the results presented in this section correspond to very simple and specific criteria, the MTBI approach is further examined in Section \ref{sec:compapproach}.

\subsection{Evaluation of the Lacanian discourses approach}
\label{sec:preliminary-lacanian-evaluation}

In this section we examine the potential of the Lacanian discourses being used to predict Real and Fake news from their \emph{headlines}.

At this initial stage, it has been decided to adopt a very simple way to quantify the presence of each possible Lacanian discourse in an enunciation. Consider the vector $\mathcal{L} = (M, A, U, H)$ where M, A, U, H stands for Master, Analyst, University and Hysteric, respectively, and may take the value 1 to indicate the presence of that type of discourse in the enunciation, and 0 otherwise. For example, $\mathcal{L} = (M, A, U, H) = (1, 0, 1, 0)$ indicates that in the corresponding enunciation traces of Master and University discourses have been identified.

The following procedure has been adopted to evaluate the Lacanian discourses approach:

\begin{itemize}
	\item[Step 1:] blind assignment of the Lacanian discourses. An expert having no access to the \emph{headlines'} labels assign $\mathcal{L} = (M, A, U, H)$ to a set of 100 \emph{headlines}.
	
	\item[Step 2:] identification of ambiguities. The expert accesses the labels of the 100 used \emph{headlines} and verifies if the same values of $\mathcal{L} = (M, A, U, H)$ have been assigned to \emph{headlines} with different labels.
	
	\item[Step 3:] non-blind re-assignment of the Lacanian discourses to get Zero ambiguities. For each of the identified ambiguities in Step 2, the expert verifies if it is possible to assign a {\bf psychoanalytically valid} alternative value of $\mathcal{L} = (M, A, U, H)$ to solve the ambiguity.
	
	\item[Step 4:] no-blind extension the Lacanian discourses assignment. Using the same criteria identified in Step 3 to solve the ambiguities the assignment of $\mathcal{L} = (M, A, U, H)$ was made for additional 200 \emph{headlines}.
\end{itemize}

Using the aforementioned procedure it was possible to find a partition of the $\mathcal{L} = (M, A, U, H)$ codes for the 300 \emph{headlines}. Table \ref{tab:partition} shows the obtained partition.

\begin{table}[htp]
\caption{$\mathcal{L} = (M, A, U, H)$ assignment for Labels equal to 0 and 1.}
\label{tab:partition}
\begin{center}
\begin{tabular}{|c|c|c|c||c|c|c|c|} \hline
\multicolumn{4}{|c|}{Label = 0} & \multicolumn{4}{||c|}{Label = 1} \\ \hline
{\bf M} & {\bf A} & {\bf U} & {\bf H} &
{\bf M} & {\bf A} & {\bf U} & {\bf H} \\ \hline
0 & 1 & 0 & 0 & 0 & 0 & 0 & 1 \\ 
0 & 1 & 0 & 1 & 0 & 0 & 1 & 0 \\
0 & 1 & 1 & 0 & 0 & 0 & 1 & 1 \\
1 & 0 & 0 & 0 & 0 & 1 & 1 & 1 \\
1 & 0 & 0 & 1 & 1 & 0 & 1 & 0 \\
1 & 1 & 0 & 0 & 1 & 0 & 1 & 1 \\
1 & 1 & 0 & 1 & 1 & 1 & 1 & 1 \\
1 & 1 & 1 & 0 &&&& \\ \hline

\end{tabular}
\end{center}
\label{default}
\end{table}%

It can be argued that the use of a non-blind reassignment of $\mathcal{L} = (M, A, U, H)$ is not fair and is just trick to get zero ambiguity. It is an understandable argument, however the procedure is justified by the following reasons:
\begin{enumerate}
	\item Psychoanalysis is not a hard science and different experts adopting different points of view may arrive at different valid conclusions. The important point to be emphasised is that a psychoanalytically valid partition of the $\mathcal{L} = (M, A, U, H)$ codes have been found.
	
	\item The length of the \emph{headlines} is very short, much shorter than a normal patient's narrative. In this case the assignment of $\mathcal{L} = (M, A, U, H)$ is much harder and a second turn of re-assignment is mandatory to arrive at useful results.
	
	\item The assignment of $\mathcal{L} = (M, A, U, H)$ to \emph{headlines} is quite different from the identification of the discourse adopted by the patient in a psychoanalytical setting where besides the narrative, body language, clothes, emotion representative gestures help the analyst to build a full representation of the patient.
\end{enumerate}

To illustrate the difficulty of assigning $\mathcal{L} = (M, A, U, H)$ consider the following \emph{headline}:
\begin{quote}
\emph{Police Turn In Badges Rather Than Incite Violence Against Standing Rock Protestors}
\end{quote}
to which has been assigned $\mathcal{L} = (M, A, U, H) = (1, 0, 1, 0)$.

The \emph{Master} assignment is justified by the mention of ``Police'' that is a figure of authority. The \emph{University} assignment is justified because this \emph{headline} is disclosing an information, some kind of knowledge. However, it could be argued that a $\mathcal{L} = (M, A, U, H) = (1, 0, 1, 1)$ assignment is better because the reference to ``Violence'' is a trace of a \emph{Hysteric} discourse. Is this assignment was adopted nothing would change because both codes are in the Label = 1 partition. On the other hand, it could also be argued that a $\mathcal{L} = (M, A, U, H) = (1, 1, 0, 0)$ is better because the verb ``Incite'' may be taken more as an opinion and interpretation than a mere disclosure of information. If this assignment were adopted, the new code belongs to the Label = 0 partition and an ambiguity would have been introduced.

This simple example shows that the assignment of Lacanian discourses is not an easy task and requires further investigation.

From Table \ref{tab:partition} a deterministic model to predict the labels can be easily derived. 

The Label should be set to Zero if $\mathcal{L} = (M, A, U, H)$ satisfies Eq. (\ref{eq:label-0}), i.e., the substitution of $\mathcal{L} = (M, A, U, H)$ into Eq. (\ref{eq:label-0}) results in 1.

\begin{equation} \label{eq:label-0}
	A.\overline{U} + M.\overline{U} + A.U.\overline{H}
\end{equation}

The Label should be set to One if $\mathcal{L} = (M, A, U, H)$ satisfies Eq. (\ref{eq:label-1}), i.e., the substitution of $\mathcal{L} = (M, A, U, H)$ into Eq. (\ref{eq:label-1}) results in 1.

\begin{equation} \label{eq:label-1}
	\overline{A}.U + \overline{M}.\overline{A}.H + U.H
\end{equation}

Equations (\ref{eq:label-0}) and (\ref{eq:label-1}) are minimum, but are not unique. They complement each other in the sense that if the substitution of the $\mathcal{L} = (M, A, U, H)$ values satisfy (\ref{eq:label-0}) they will not satisfy (\ref{eq:label-1}), and vice-versa.

Table \ref{tab:partition} and Equations (\ref{eq:label-0}) and (\ref{eq:label-1}) suggest that the presence of the Analyst discourse, $\mathcal{L} = (M, A, U, H) = (*, 1, *, *)$ almost always means that the corresponding \emph{headline} refers to a Real news. On the other hand, the presence of the University discourse, $\mathcal{L} = (M, A, U, H) = (*, *, 1, *)$ almost always means that the corresponding \emph{headline} refers to a Fake news.

Finally the assignment of $\mathcal{L} = (M, A, U, H)$ values was extended to another 300 \emph{headlines} to allow further evaluation of the procedure in Section \ref{sec:compapproach}.

\section{Computational Approach}
\label{sec:compapproach}
This section describes the Machine Learning (ML) experiments of our approach to classify and predict whether a \emph{headline} is Fake or Real. The following subsections describe the various adopted procedures and showcase the results.


\subsection{Algorithmic Personality Type (MTBI) Assignment to the Dataset's Headlines}
\label{sec:mbti_ml}
Several open-source codes and projects have been developed to extract the personality types from a text as, for example, the works available in \cite{mehul2020} and \cite{saini2020}. 

In this work, the dataset file was submitted to the personality traits assignment algorithm developed by \cite{mehul2020}, which is implemented in the programming language Python\footnote{https://www.python.org/}. The analysis is performed using linguistic cues like word repetition, number of verbs, nouns, detection of emotion in text and so on. We are mainly interested in the application of such a theory and not in the implementation of the theory itself as it has already been done and evaluated by other researchers as in \cite{Iskandar2021}. Some of the approaches are able to determine types like E/I, S/N and T/F with at least 90\% accuracy \cite{Li2018} and, more recently, the J/P type with 81\% and 65\% accuracy, depending on the used dataset \cite{Choong2021}.

The MBTI algorithm takes as input a text and gives as output 4 groups of variables (Extroversion/Introversion, Thinking/Feeling and so on) in the range of [0,1], which represent the dimensions described in Section \ref{sec:personalityapproach}, and the binary classification task of the ML models is to classify the \emph{headlines} whether they are Real (dataset label is 0) or Fake (dataset label is 1).

The mean of each personality type group is presented in Table \ref{tab:mean_types}.

\begin{table}[!htb] 
\caption{Mean, rounded to 2 decimals,  of each variable (type) obtained from the algorithm}
\label{tab:mean_types}
\begin{center}
\begin{tabular}{||c c||} 
    \hline
Personality Type & Mean \\ [0.5ex] 
    \hline\hline
Introversion - Extroversion & 0.45 \& 0.55\\ 
    \hline
Sensing - Intuiting& 0.43 \& 0.57 \\
    \hline
Thinking - Feeling & 0.33 \& 0.67 \\
    \hline
Judging - Perceiving & 0.49 \& 0.51\\
\hline
\end{tabular}
\end{center}
\end{table}

\FloatBarrier

\begin{figure*}[!htb] 
   \begin{center}
      \includegraphics[keepaspectratio=true, width=0.75\textwidth]{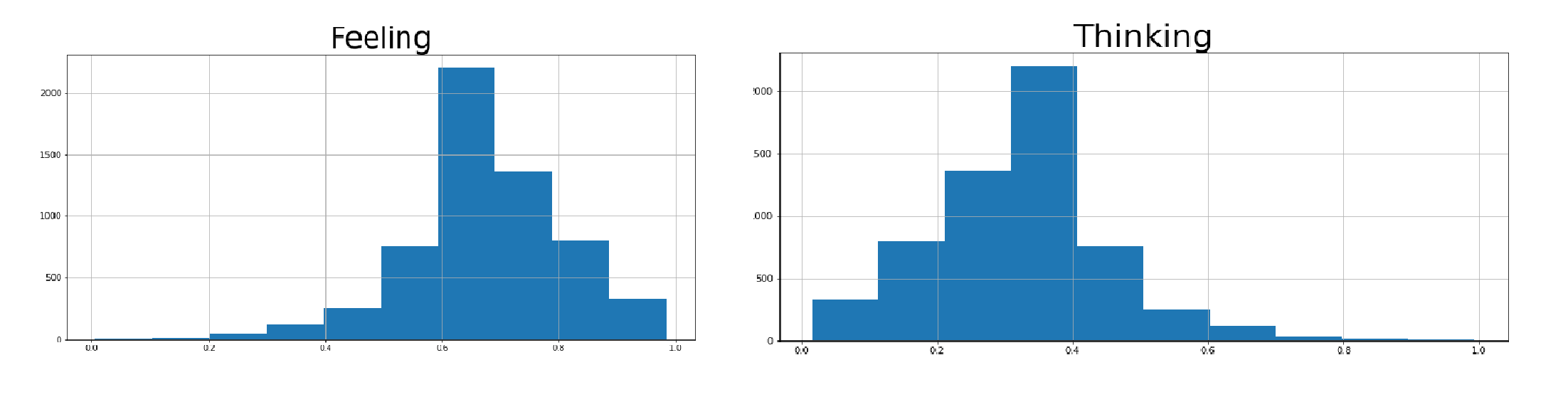}
   \end{center}
    \caption{Distribution of types Feeling (mean 0.67) and Thinking (mean 0.33) }
    \label{fig:fee-think}
\end{figure*}

\begin{figure*}[!htb] 
   \begin{center}
      \includegraphics[keepaspectratio=true, width=0.75\textwidth]{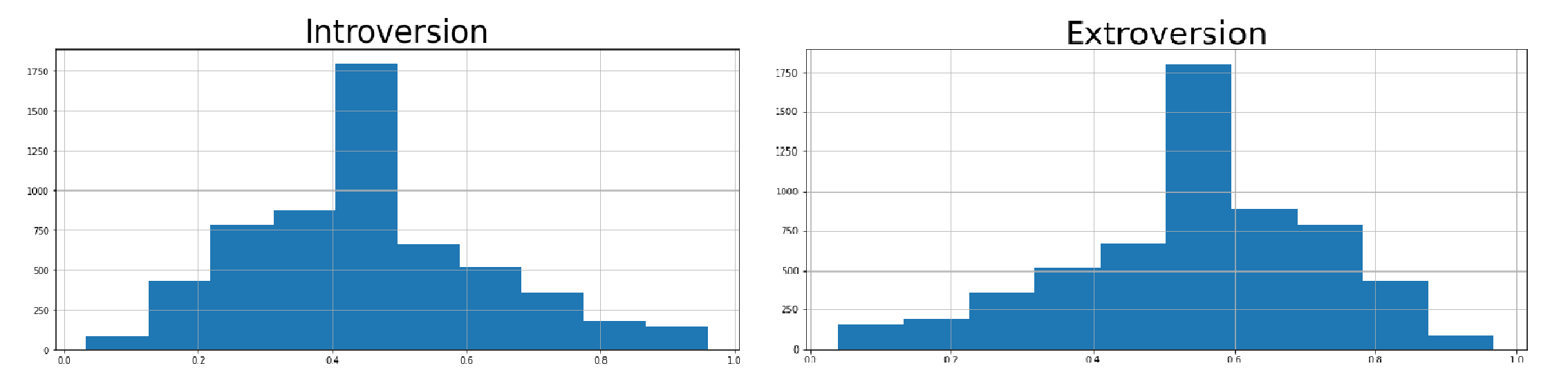}
   \end{center}
    \caption{ Distribution of types Introversion (mean 0.49) and Extroversion (mean 0.55)}
    \label{fig:intro-extr}
\end{figure*}

\begin{figure*}[!htb] 
   \begin{center}
      \includegraphics[keepaspectratio=true, width=0.75\textwidth]{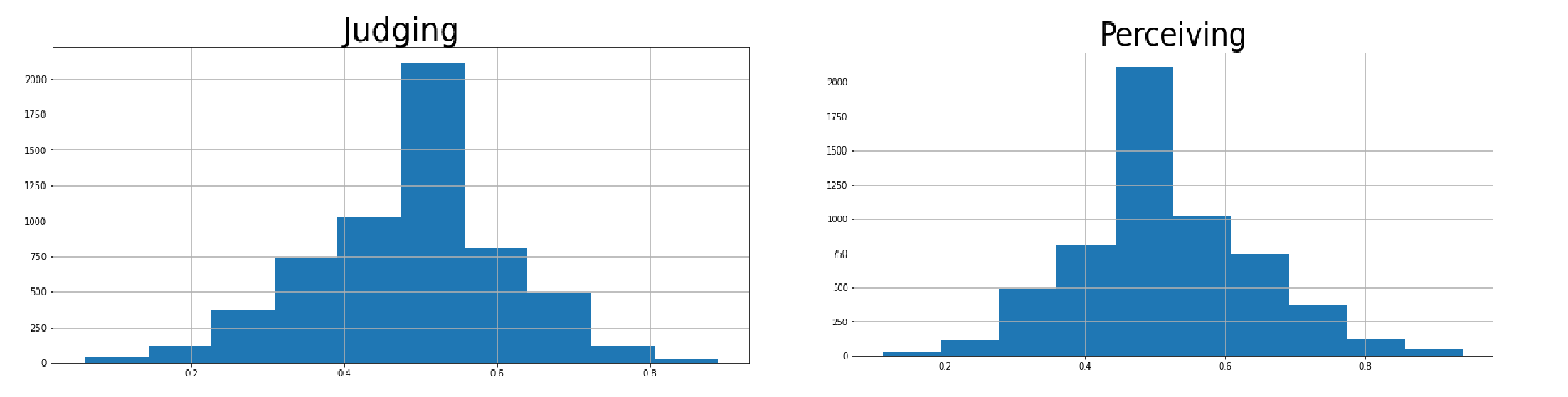}
   \end{center}
    \caption{ Distribution of types Judging (mean 0.45) and Perceiving (mean 0.51)}
    \label{fig:jdg-perc}
\end{figure*}

\begin{figure*}[!htb] 
   \begin{center}
      \includegraphics[keepaspectratio=true, width=0.75\textwidth]{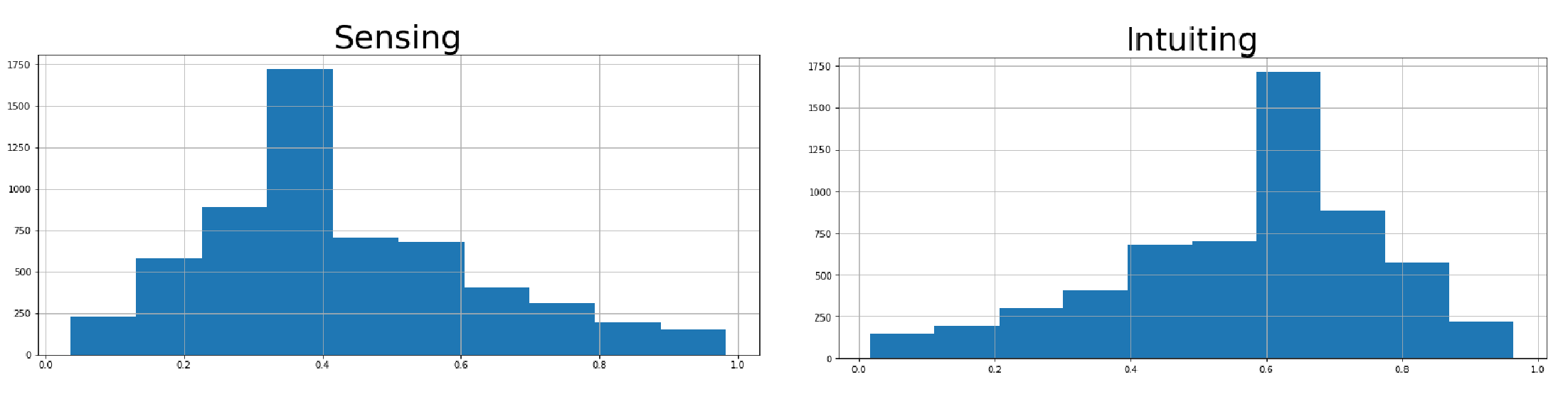}
   \end{center}
    \caption{ Distribution of types Sensing (mean 0.43) and Intuiting (mean 0.57)}
    \label{fig:send-intu}
\end{figure*}

The distribution of the obtained types is illustrated in Figures \ref{fig:fee-think} (Thinking - Feeling), \ref{fig:intro-extr} (Introversion - Extroversion), \ref{fig:jdg-perc} (Judging - Perceiving) and \ref{fig:send-intu} (Sensing - Intuiting). 

\subsection{MTBI Fake or Real news prediction using ML}
\label{sec:fake_real_ml}
We selected four different ML algorithms, namely: i) Support Vector Classifier (SVC), ii) LightGBM, iii) Extra-Trees-Classifier and iv) GaussianNB (GNB) for comparison and tested them with the following groups of independent personality types in an attempt to identify the impact of different personality types in the field of Fake/Real news detection:

 \begin{itemize}
 	\item Type 1: Introversion, Intuition, Thinking, Judging (INTJ)
	\item Type 2: Extroversion, Sensing, Feeling, Perceiving  (ESFP)
	\item Type 3: Introversion, Sensing, Thinking, Perceiving (ISTP)
 \end{itemize}
 
Each algorithm is run using default parameters, is trained with 25\% of data, and the results are presented in Table \ref{tab:models_acc}. It can be seen that Types 1, 2 led to 59\% accuracy by the LightGBM algorithm, while Type 3 had almost always the lowest score. The GaussianNB had the worst performance among all. It is noticeable that hyperparameter tuning didn't improve the accuracy and didn't provide any additional insight about the characteristics of the task and dataset.

\begin{table} [!htb]
\caption{The summary of the performance of the ML algorithms and the relative achieved balanced accuracy}
\label{tab:models_acc}
\begin{center}
\begin{tabular}{||c c c c||} 
 \hline
 {\bf Model} & {\bf Type 1} & {\bf Type 2} & {\bf Type 3} \\ [0.5ex] 
 \hline
 Extra-Trees-Classifier & 58\% & 57\% & 56\% \\ 
 \hline
 LightGBM & 59\% & 59\% & 58\% \\
 \hline
 SVC & 57\% & 57\% & 55\% \\
 \hline
 GaussianNB & 54\% & 54\% & 55\% \\ [1ex] 
 \hline
\end{tabular}
\end{center}
\end{table}

\begin{figure}[!htb]
    \centering
    \includegraphics[width=0.45\textwidth]{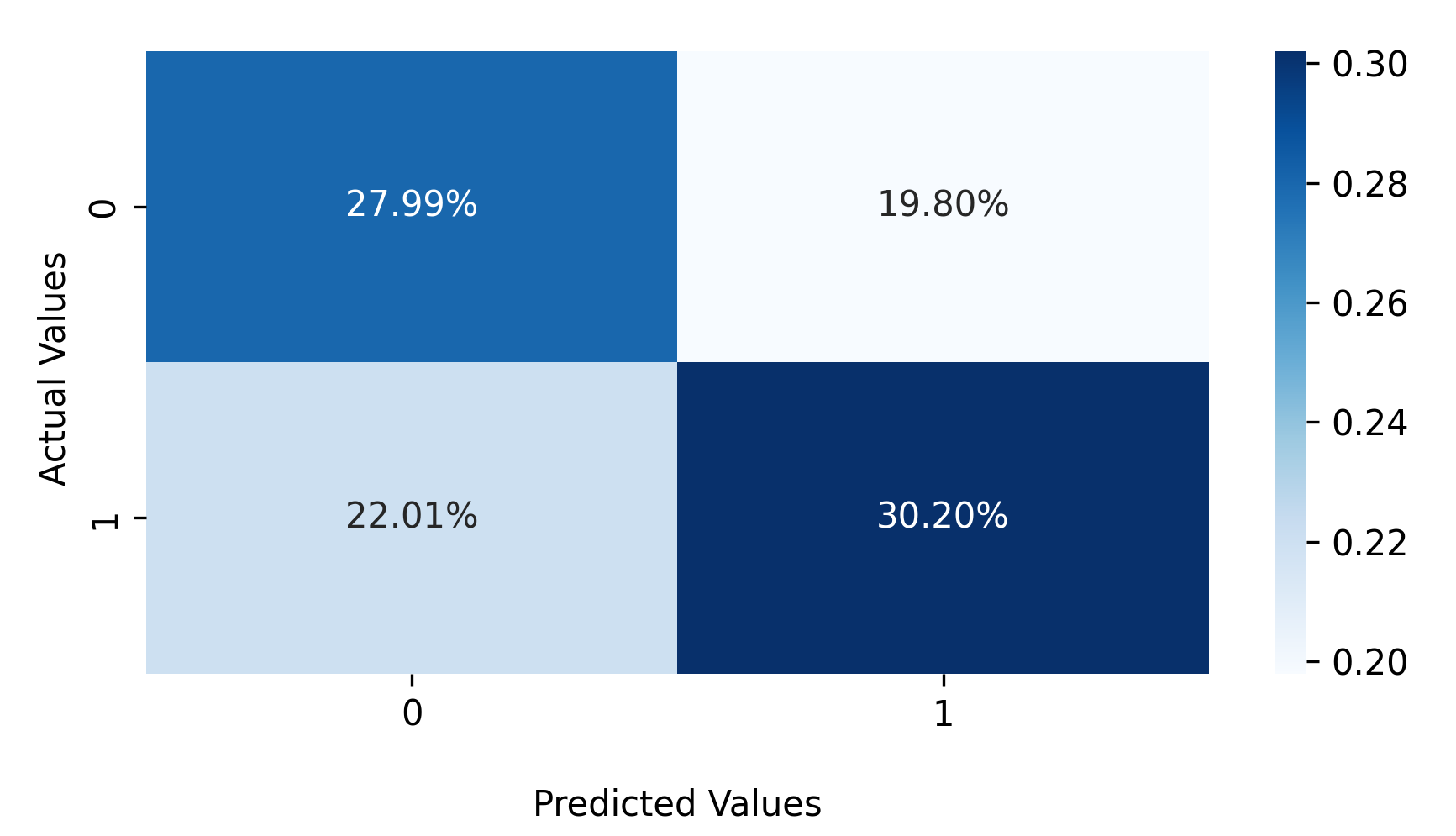}
    \caption{Confusion Matrix of the LightGBM algorithm  for the Type 1 experiment}
    \label{fig:cm_1}
\end{figure}

The confusion matrix derived from the LightGB algorithm, Type 1 experiment, is shown in Figure \ref{fig:cm_1}. It can be seen that the best-scoring model is able to correctly predict a fake news \emph{headline} as Fake 30.2\% of the time (bottom right corner of figure), while there is a probability of approximately 22\% to categorize a fake news \emph{headline} as Real (bottom left). Regarding the real news label, approximately 27.9\% correctly categorize correctly a real news \emph{headline} as Real (top left) and 19.8\% correctly to categorize a real news headlines as Fake (top right).

\FloatBarrier

The obtained results confirm what has already been stated in Section \ref{sec:preliminary-mtbi-evaluation} that MBTI is  not the ideal tool for predicting \emph{headline} labels, as none of the ML algorithms achieved an accuracy score much better than a random prediction.

\subsection{Lacanian Discourses Fake or Real news prediction using ML}
\label{sec:lacan_ml}
This is a two steps procedure:
 \begin{itemize}
 	\item Step 1: Assign the Lacanian Discourses to the \emph{headlines} either manually or automatically.
	\item Step 2: Using the results from Step 1, predict the labels of the \emph{headlines}, as Fake or Real.
 \end{itemize}

This gives rise to four possible combinations of discourses assignment and labels prediction:

 \begin{itemize}
 	\item \emph{Manually} assign the Lacanian Discourses and predict labels using the deterministic model discussed in Section \ref{sec:preliminary-lacanian-evaluation} (Section \ref{sec:man_lacan_1}).
 	\item \emph{Manually} assign the Lacanian Discourses and predict labels using ML algorithms (Section \ref{sec:man_lacan_2}).
 	\item \emph{Automatically} assign the Lacanian Discourses  using the language model GPT-3 and predict labels using the deterministic model discussed in section \ref{sec:preliminary-lacanian-evaluation} (Section \ref{sec:auto_lacan_1}).
 	\item \emph{Automatically} assign the Lacanian Discourses  using the language model GPT-3 and predict labels using ML algorithms (Section \ref{sec:auto_lacan_2}).
 \end{itemize}
 
For the first two approaches, we randomly selected 600 \emph{headlines} and manually assigned the corresponding Lacanian discourses. 

In the case of the latter two approaches, we used the aforementioned 600 manually assigned \emph{headlines} to train the language model GPT-3 (Generative Pre-trained Transformer 3) in an attempt to achieve automatic Lacanian Discourses assignment to solve the problem of \emph{Step 1}. 

GPT-3 is a neural network machine learning model, developed by OpenAI, trained using internet data to generate any type of text that enables developers to train and deploy AI models. It provides a variety of tools and services for data preprocessing and model training, and its capabilities include but are not limited to: i) generating, classifying, translating and summarising text; ii) generating and answering questions; iii) generating images and audio from text \cite{Dale2020} and can be accessed and used on OpeanAI's main website \footnote{\url{https://openai.com/api/}}.

\subsubsection{Manual Lacanian discourses assignment and deterministic model labels prediction (no ML used)}
\label{sec:man_lacan_1}
This case was introduced and implemented in Section \ref{sec:preliminary-lacanian-evaluation}. The accuracy is almost 100\%, however, it is not a practical alternative because the  manual assignment of Lacanian discourse is only feasible for a very small number of \emph{headlines}. Nevertheless, it is an important theoretical case because it has shown the existence of a psychoanalytic valid partition of the codes.

\subsubsection{Manual Lacanian discourses assignment and ML labels prediction}
\label{sec:man_lacan_2}

The ML algorithms presented in Section \ref{sec:fake_real_ml} were trained and evaluated with the 600 manually Lacanian discourses assigned to the \emph{headlines}. The accuracy of the labels prediction is summarised in Table \ref{tab:models_acc_2}.

\begin{table}[!htb]
\caption{Performance of the ML algorithms and the relative achieved balanced accuracy of labels prediction for manually assigned Lacanian discourses}
\label{tab:models_acc_2}
\begin{center}
\begin{tabular}{||c c||} 
 \hline
 {\bf Model} & {\bf Accuracy} \\ [0.5ex] 
 \hline
 Extra-Trees-Classifier & 97\% \\ 
 \hline
 LightGBM & 97\% \\
 \hline
 SVM & 97\% \\
 \hline
 GaussianNB & 92\% \\ [1ex] 
 \hline
\end{tabular}
\end{center}
\end{table}

\FloatBarrier

As expected, the accuracy is slightly decreased compared to the first approach. It is important to realise that this is not a typical ML scenario as the dataset is very small and the models may be overfitting: the model fits perfectly against its training data and thus achieves very high accuracy. 

Once again we emphasise that the manual assignment of Lacanian discourses is rather impractical but it confirms the potential of the psychoanalytic-based method.

\subsubsection{Automatic Lacanian discourses assignment and deterministic model labels prediction}
\label{sec:auto_lacan_1}

OpenAI\footnote{https://beta.openai.com/docs/models/gpt-3} provides an API to work with four different models: Davinci, Curie, Babbage, and Ada. Each one of them has different capabilities, advantages and disadvantages, and after some experimenting we decided to work with the model Ada because of its speed and its capabilities more adapted to the nature of the classification under study. 

We first created a set of classification Ada models using 100, 200, and 500 out of 600 manually Lacanian discourses assigned \emph{headlines} to evaluate its accuracy in predicting the discourses of the last 100 \emph{headlines}. 

\begin{figure}[!htb]
    \centering
    \includegraphics[width=0.45\textwidth]{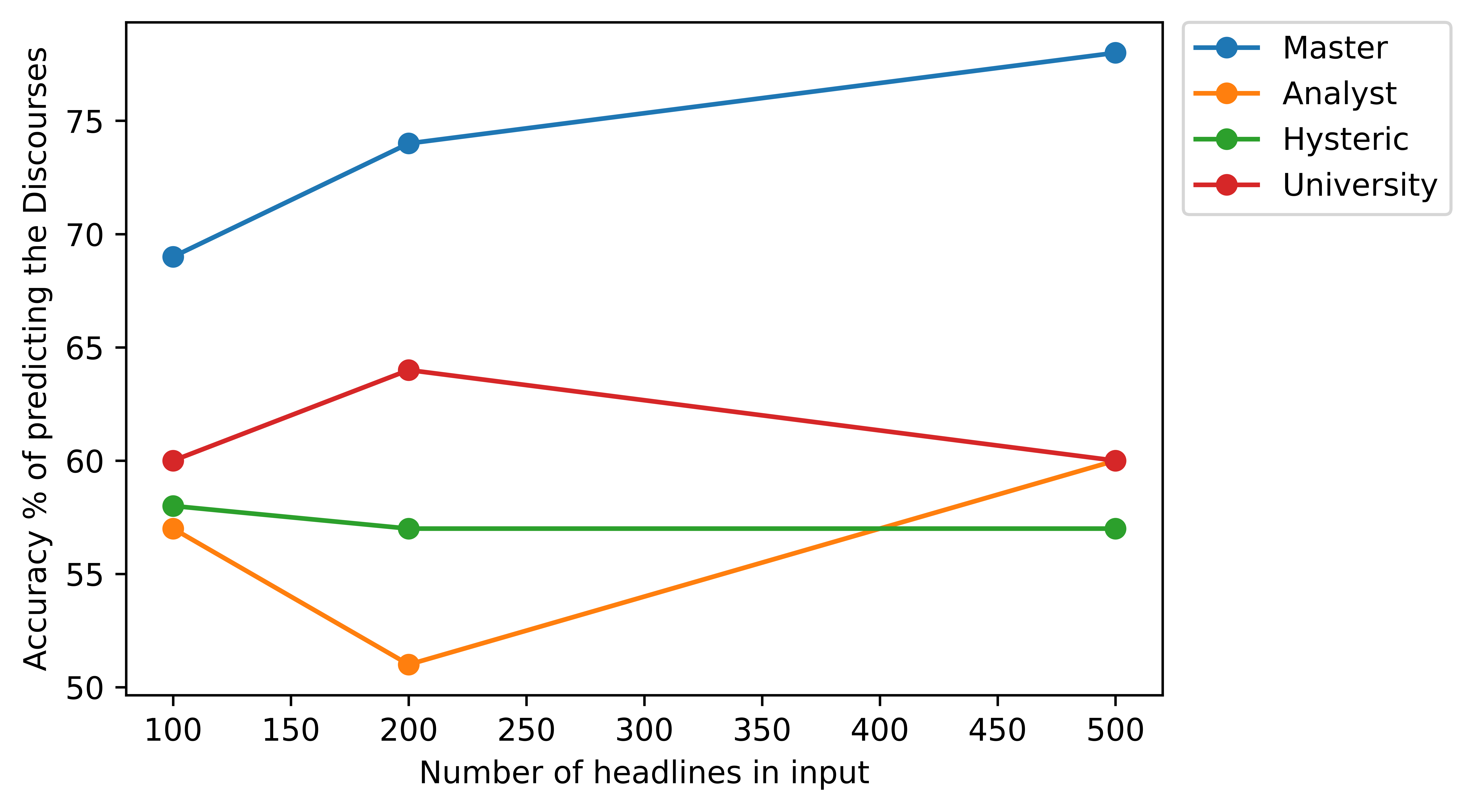}
 \caption{The X axis represents the number of manually assigned \emph{headlines} used to train the classification model; the Y axis represents the accuracy of predicting the discourses of the last 100 manually assigned \emph{headlines}.}
 \label{fig:step_1_1}
\end{figure}

The results of this experiment are shown in Figure \ref{fig:step_1_1}. It can be seen that  as the model is getting trained with more assigned \emph{headlines}, it gets better at classifying and labeling the Master discourse, while it gets worse at labeling the Analyst discourse and a bit better at labeling the University discourse when it is trained with 200 headlines, but it ends up at the same accuracy of 100 assigned \emph{headlines} when the input is 500 assigned \emph{headlines}.

\FloatBarrier

\begin{figure}[!htb]
    \centering
    \includegraphics[width=0.45\textwidth]{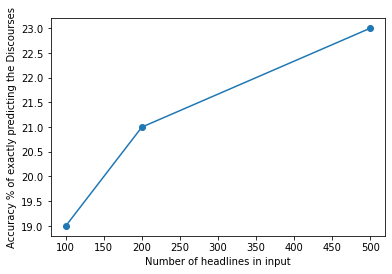}
 \caption{The X axis represents the number of manually assigned \emph{headlines} used to train the classification model; the Y axis represents the accuracy of exactly predicting the four discourses of the last 100 manually assigned \emph{headlines}.}
 \label{fig:step_1_2_1}
\end{figure}

The overall accuracy of exactly predicting exactly the four discourses of the last 100 \emph{headlines} is illustrated in Figure \ref{fig:step_1_2_1}. It can be seen that the accuracy is below 30\% but with a slight increase as the number of assigned discourses increases.

\FloatBarrier

Giving the 600 manually assigned \emph{headlines} as training in the Ada model and evaluating the whole dataset using the deterministic model described in Section \ref{sec:preliminary-lacanian-evaluation}, the labels prediction accuracy is only 66\%. This is a poor result, however much better than what was obtained with the MTBI approach, due to a poor automatic discourses assignment accuracy, as discussed before.

\subsubsection{Automatic Lacanian discourses assignment and ML labels prediction}
\label{sec:auto_lacan_2}

Following the same procedure as described in Section \ref{sec:auto_lacan_1}, using 100, 300, and 600 as training points, the results of the ML models labels prediction of the \emph{headlines} are shown in Table \ref{tab:models_acc_openai_1}. The Extra-Trees-Classifier, LightGBM, SVC, and GaussianNB models best accuracy is 66\%. 

\begin{table}[!htb]
\caption{The ML models' performance and the relative achieved balanced labels prediction accuracy}
\label{tab:models_acc_openai_1}
\begin{center}
\begin{tabular}{||c c c c||} 
 \hline
 {\bf Model} & \multicolumn{3}{c||}{\bf{Number of training points}}\\ 
 & {\bf 100} & {\bf 300} & {\bf 600} \\ [0.5ex] 
 \hline
 Extra-Trees-Classifier & 63\% & 64\% & 66\% \\ 
 \hline
 LightGBM & 63\% & 64\% & 66\% \\
 \hline
 SVC & 63\% & 64\% & 66\% \\
 \hline
 GaussianNB & 62\% & 63\% & 65\% \\ [1ex] 
 \hline
\end{tabular}
\end{center}
\end{table}

\FloatBarrier

\begin{figure}[!htb]
    \centering
    \includegraphics[width=0.45\textwidth]{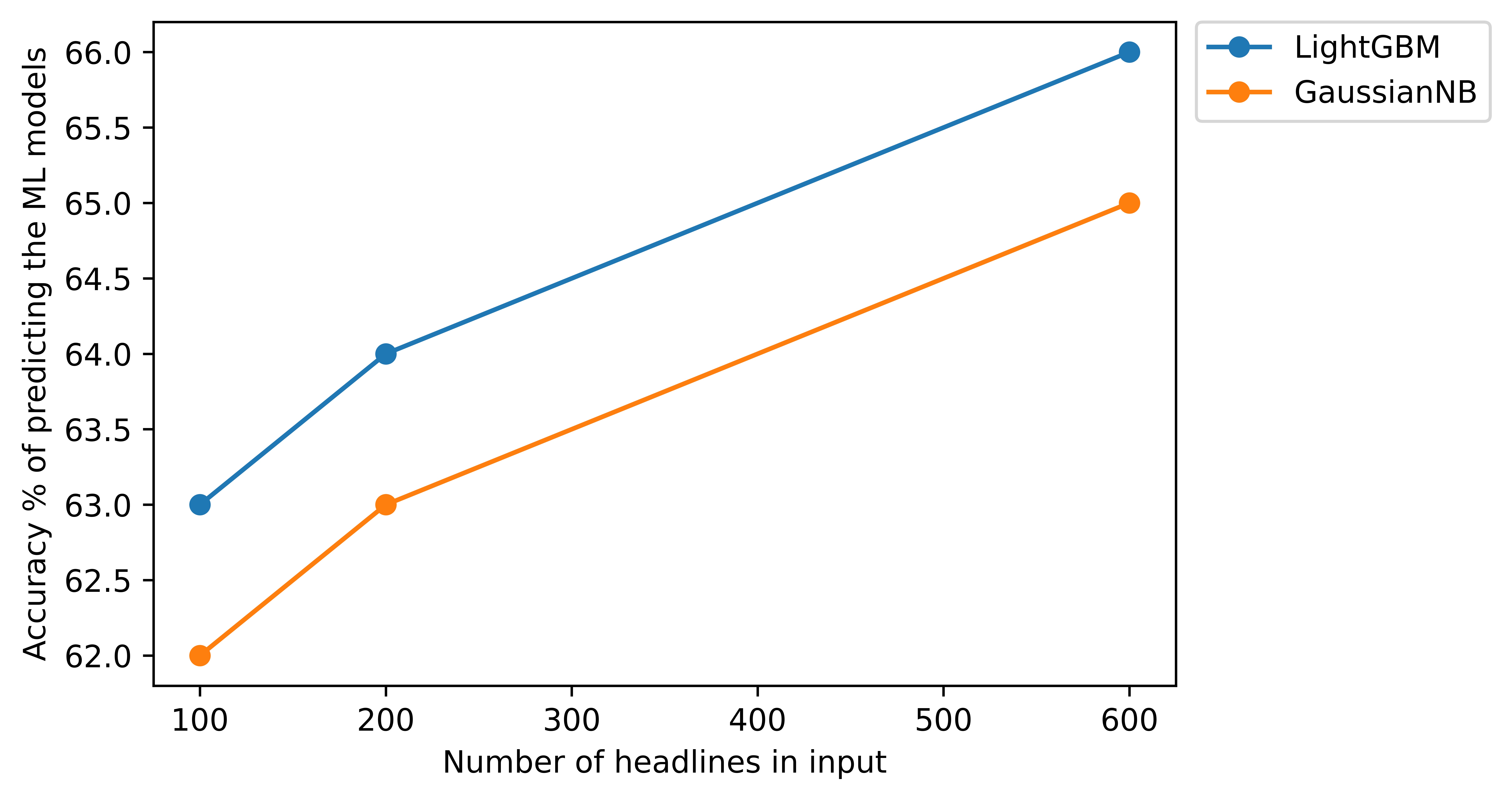}
 \caption{The X axis represents the number of manually assigned \emph{headlines} used to train the classification model; the Y axis represents the ML models dataset labels prediction accuracy.}
 \label{fig:models_openai_ml_11}
\end{figure}

The accuracy gets better as the number of training points increases, as shown in Figure \ref{fig:models_openai_ml_11}.

To estimate how many headlines should be used for training the models to achieve a 70\% accuracy threshold we did a log-fitting of the obtained points using Eq. (\ref{eq:log_fit_1}):

\begin{equation} \label{eq:log_fit_1}
    a \log (b x+1)+0.5
\end{equation}.

\begin{figure}[!htb]
    \centering
    \includegraphics[width=0.45\textwidth]{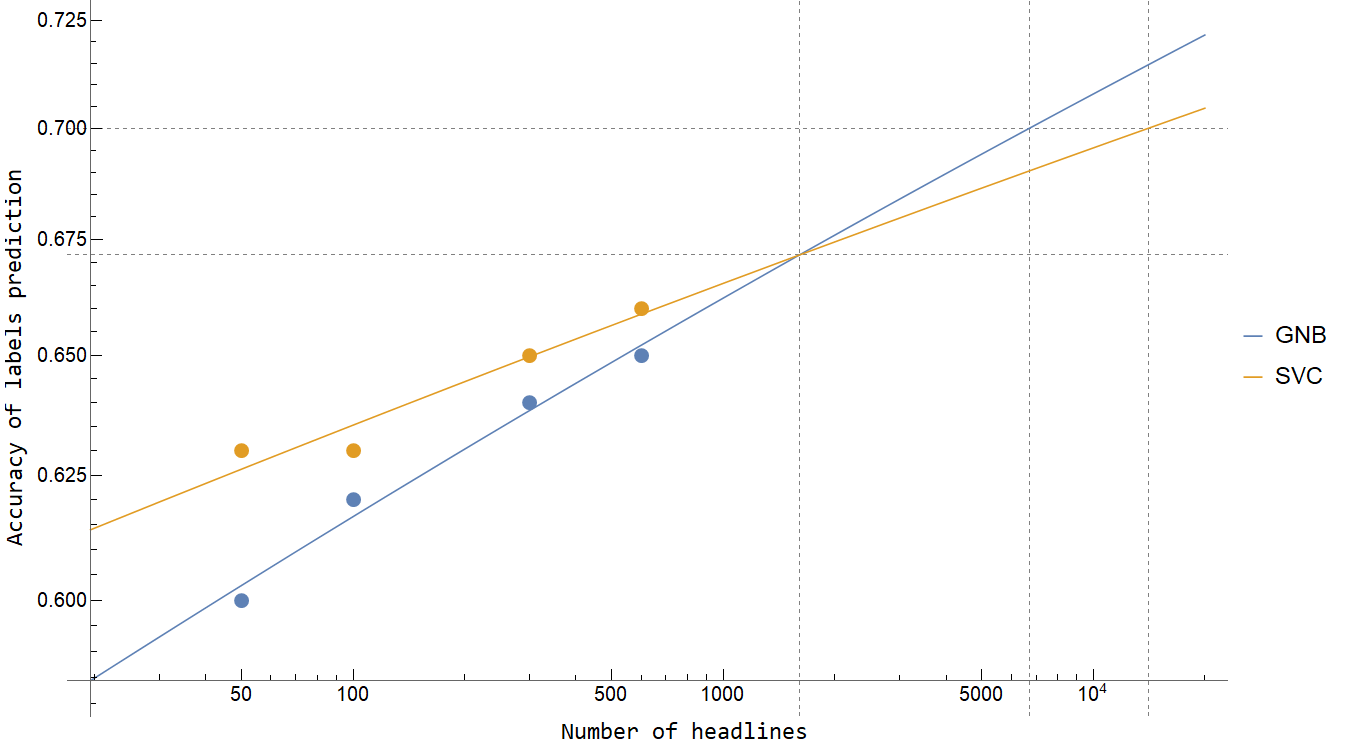}
 \caption{The X axis represents the number of manually assigned \emph{headlines} used to train the classification model; the Y axis the dataset labels prediction accuracy of the GNB (blue), SVC (yellow) ML models.}
 \label{fig:models_openai_ml_22}
\end{figure}

Figure  \ref{fig:models_openai_ml_22}, that is in logarithmic scale, shows the results. 

\FloatBarrier

The parameters for the log function are:
\begin{itemize}
    \item[--] GNB: a = 0.0198352, b = 3.56483;
    \item[--] SVC: a = 0.0130760, b = 312.088.
\end{itemize}

The number of points needed to a 70\% accuracy threshold is 6700 for the GNB model, and 14000 for the SVC model. With 1600 points the two models provide the same 67\% accuracy, and after that point the GNB model performs better. 

 Despite the power of models provided by the OpenAI the problem of automatically assign the Lacanian discourses remains to be solved to benefit from the great potential of psychoanalytic-based approach to be used to extract information from enunciations.

\section{Conclusions}
\label{sec:conclusions}
Different approaches of combining psychological and social dimensions with computational methods which include computational psycholinguistics, personality traits, behavioural analysis, emotional states and cognitive psychology methods have been used to derive information from user-related and generated data. However, in various contexts where user perception is more susceptible to emotional or some other form of bias, the reliability of user data is often compromised. 

 Compared to all those approaches, the approach proposed in this work is fundamentally different since we adopt a psychoanalytic perspective; in particular, we employ the powerful notion of Lacanian discourse types. To the best of our knowledge, this is the first attempt to systematically bring together psychoanalysis and computing. 
 
 From the experimental point of view, as a preliminary case study scenario, we looked at the problem of identifying from the \emph{headlines} of published news if they are Real or Fake.
 
 To start with, we evaluated the psychological-based approach of deriving the Real or Fake characteristic from the personality traits associated to each \emph{headline}. In Section \ref{sec:preliminaryevaluation}, it has been shown that this approach is not better than a random prediction. This conclusion has been confirmed in Section \ref{sec:fake_real_ml}, where different ML algorithms achieved at most 56\% accuracy in label prediction.
 
 Then, we evaluated the psychoanalytic-based approach. It is a two steps procedure: i) each \emph{headline} received a Lacanian discourses assignment $\mathcal{L} = (M, A, U, H)$, where M, A, U, H stands for Master, Analyst, University and Hysteric, respectively, and may take the value 1 to indicate the presence of that type of discourse in the enunciation, and 0 otherwise.; ii) Real or Fake label prediction from the $\mathcal{L} = (M, A, U, H)$.
 
 A partition of the $\mathcal{L} = (M, A, U, H)$ between Real and Fake labels have been obtained and a deterministic model for label prediction has been derived. This indicates the possibility of achieving a theoretical 100\% accuracy of label prediction. However, the described procedure is not practical as the Lacanian discourses assignment was manually done. The results were confirmed in Section \ref{sec:lacan_ml}. 
 
 A first attempt to automatically assign the Lacanian discourses was done using the environment provided by OpenAI. The label prediction accuracy decreased to 66\%.
 
 In summary, we conclude that the psychoanalytic-based approach has a great potential to extract information from user-generated data.
 
 As future work, an automatic procedure to assign the Lacanian discourses will be developed, and the technique will be applied to other problems rather than the Real / Fake news detection. The procedure will be extended to non-binary classification problems and to multimedia user-generated data.

\bibliographystyle{IEEEtran}
\bibliography{references.bib}

\end{document}